\documentclass{bmvc2k}
\usepackage{amsmath,amssymb,amsfonts}
\usepackage{adjustbox}
\usepackage{multicol}
\usepackage{multirow}
\usepackage{booktabs}
\usepackage{hyperref}
\usepackage{colortbl}
\usepackage{pifont}
\usepackage[capitalize]{cleveref}

\newcommand\blfootnote[1]{%
  \begingroup
  \renewcommand\thefootnote{}\footnote{#1}%
  \addtocounter{footnote}{-1}%
  \endgroup
}

\title{CMAMRNet: A Contextual Mask-Aware Network Enhancing Mural Restoration Through Comprehensive Mask Guidance}

\addauthor{Yingtie Lei}{leiyingtie@outlook.com}{1}
\addauthor{Fanghai Yi}{3122005058@mail2.gdut.edu.cn}{2}
\addauthor{Yihang Dong}{dongyihang23@mails.ucas.ac.cn}{3}
\addauthor{Weihuang Liu}{yc37498@um.edu.mo}{1}
\addauthor{Xiaofeng Zhang}{framebreak@sjtu.edu.cn}{4}
\addauthor{Zimeng Li}{li_zimeng@szpu.edu.cn}{5}
\addauthor{Chi-Man Pun}{cmpun@umac.mo}{1}
\addauthor{Xuhang Chen}{https://cxh.netlify.app}{1$^\ast$}

\addinstitution{
 University of Macau
}
\addinstitution{
 Guangdong University of Technology
}
\addinstitution{
 University of Chinese Academy of Sciences
}
\addinstitution{
 Shanghai Jiao Tong University
}
\addinstitution{
 Shenzhen Polytechnic University
}

\runninghead{Lei, et al}{CMAMRNet}

\begin{document}

\maketitle
\blfootnote{$^\ast$ Corresponging author.}
\begin{abstract}
Murals, as invaluable cultural artifacts, face continuous deterioration from environmental factors and human activities. Digital restoration of murals faces unique challenges due to their complex degradation patterns and the critical need to preserve artistic authenticity. Existing learning-based methods struggle with maintaining consistent mask guidance throughout their networks, leading to insufficient focus on damaged regions and compromised restoration quality. We propose CMAMRNet, a Contextual Mask-Aware Mural Restoration Network that addresses these limitations through comprehensive mask guidance and multi-scale feature extraction. Our framework introduces two key components: (1) the Mask-Aware Up/Down-Sampler (MAUDS), which ensures consistent mask sensitivity across resolution scales through dedicated channel-wise feature selection and mask-guided feature fusion; and (2) the Co-Feature Aggregator (CFA), operating at both the highest and lowest resolutions to extract complementary features for capturing fine textures and global structures in degraded regions. Experimental results on benchmark datasets demonstrate that CMAMRNet outperforms state-of-the-art methods, effectively preserving both structural integrity and artistic details in restored murals. The code is available at~\href{https://github.com/CXH-Research/CMAMRNet}{https://github.com/CXH-Research/CMAMRNet}.
\end{abstract}

\section{Introduction}
\label{sec:intro}
Murals, as irreplaceable cultural artifacts, face accelerating deterioration due to environmental factors and human activities~\cite{wang2023current}. While traditional restoration through mechanical and chemical methods~\cite{bomin2018scientific} requires extensive expertise, digital restoration has emerged as a promising alternative~\cite{mol2021digital}, offering safe, reversible and efficient solutions that preserve the authenticity of the mural without physical contact. Among various digital restoration techniques, image inpainting stands out as a promising approach.

The rapid development of artificial intelligence and deep learning has provided potential technical solutions for image inpainting~\cite{bai2024retinexmamba,bai2025lensnet,xia2025dlen}. Early CNN-based approaches pioneered irregular mask handling through partial convolutions~\cite{yu2019free} and U-Net fusion~\cite{hong2019deep}. Advanced techniques introduced Contextual Residual Aggregation~\cite{yi2020contextual} and Region Normalization~\cite{yu2020region} for feature refinement. Recent frameworks enhanced performance through multi-level Siamese filtering (MISF~\cite{li2022misf}), Fast Fourier Convolutions (LaMa~\cite{suvorov2022resolution}), and attentional high-resolution processing (CoordFill~\cite{liu2023coordfill}). Transformer-based methods like T-Former~\cite{deng2022t} and SyFormer~\cite{wu2024syformer} improved long-range dependency modeling through linear attention and structural priors, while DWTNet~\cite{shamsolmoali2024distance} and HINT~\cite{chen2024hint} introduced distance-aware transformers and hierarchical feature interaction for adaptive context modeling.
However, applying these deep learning methods for mural restoration poses unique challenges: (1) murals have unique artistic styles and complex degradation patterns, such as extensive damage, intricate cracks, and environmental wear; (2) mural restoration requires precise conservation of artistic and historical elements, exceeding the demands of natural image inpainting; (3) limited intact mural images restrict available training data, impairing network learning of restoration patterns. These challenges necessitate specialized inpainting approaches for effective mural restoration.

Specialized for mural restoration, datasets like MuralDH~\cite{xu2024comprehensive} have enabled significant advances. Recent methods explored various approaches: multiscale adaptive partial convolutions with stroke-like mask simulation~\cite{wang2021thanka}, line drawing structural guidance~\cite{li2022line}, DenseNet-based color authenticity preservation~\cite{xu2023deep}, and edge-guided diffusion techniques for natural crack repair~\cite{xu2024muraldiff}. However, these methods have notable limitations: earlier approaches~\cite{wang2021thanka,li2022line,xu2023deep} struggle with global context and fine-grained preservation, while recent diffusion-based methods may generate historically inaccurate details and face cross-mural adaptation challenges due to complex fine-tuning requirements. Moreover, existing methods inadequately maintain mask guidance throughout their networks, as mask information weakens through layers without explicit maintenance mechanisms, resulting in insufficient sensitivity to degraded regions requiring restoration. This limited mask awareness prevents models from maintaining a precise focus on the degraded areas that require restoration. 

\begin{figure}[t]
    \centering
    \includegraphics[width=.8\linewidth]{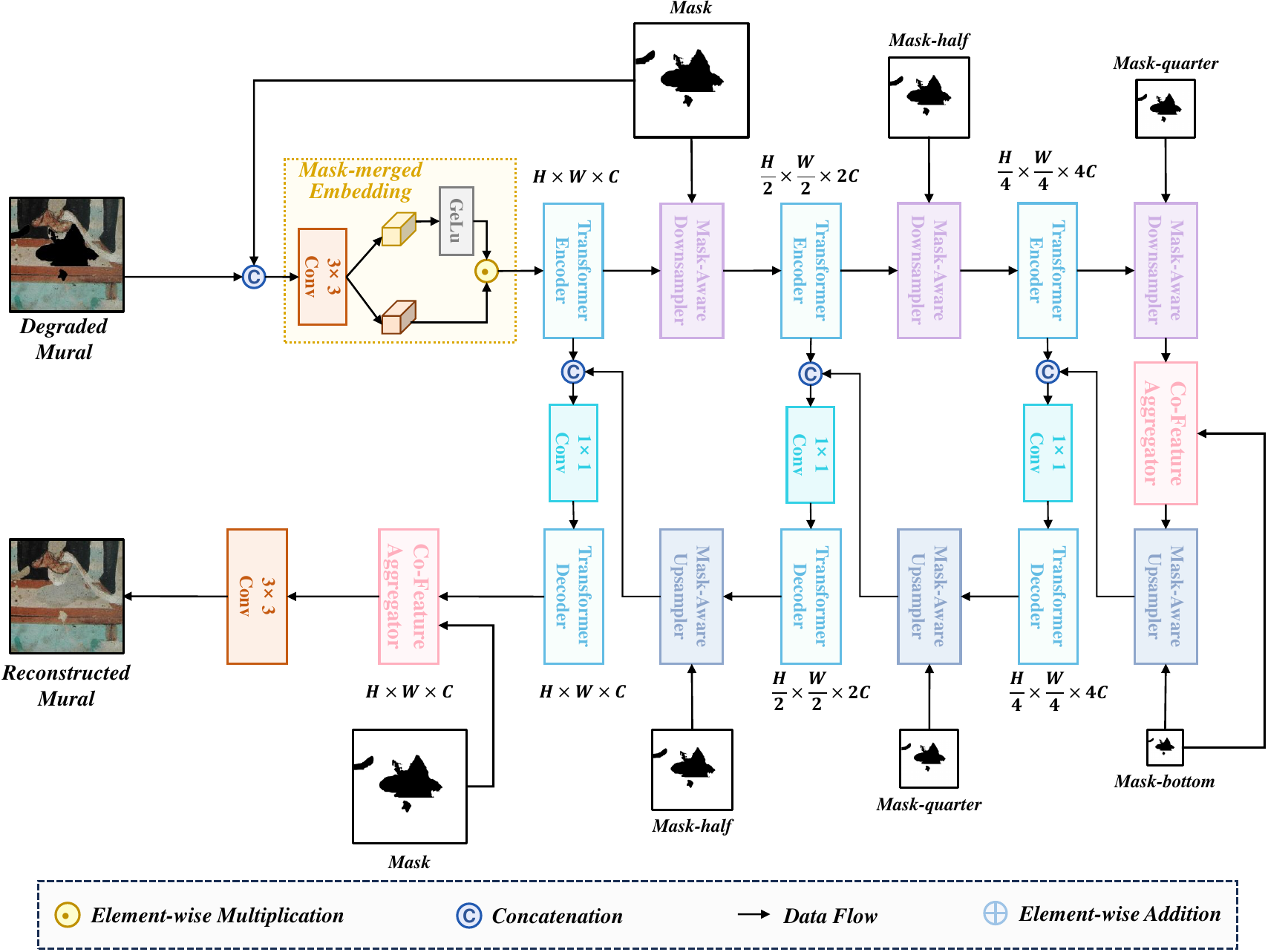}
    \caption{Architecture overview of the proposed CMAMRNet. The framework is built upon a U-shaped backbone with Transformer-based encoders and decoders for hierarchical feature extraction. It introduces two key innovations: (1) Mask-Aware Up/Down-Sampler (MAUDS) modules that preserve mask information across resolution changes via selective channel processing and mask-guided feature integration, and (2) Co-Feature Aggregator (CFA) modules operating at both highest and lowest resolutions to capture comprehensive features of fine details and global structures for holistic restoration quality.
    } 
    \label{fig:CMAMRNet}
\end{figure}
To address these limitations, we propose \textbf{C}ontextual \textbf{M}ask-\textbf{A}ware \textbf{M}ural \textbf{R}estoration \textbf{N}etwork (\textbf{CMAMRNet}), a novel framework that maintains comprehensive mask guidance throughout its processing pipeline while capturing rich contextual information at multiple scales. Our architecture leverages Transformer modules for effective long-range dependency modeling, crucial for preserving global structural coherence in mural restoration. The framework introduces two key components: (1) the Mask-Aware Up/Down-Sampler (MAUDS) module that ensures consistent mask sensitivity across resolution scales through dedicated channel-wise feature selection and mask-guided feature fusion, and (2) the Co-Feature Aggregator (CFA) that operates at both highest and lowest resolutions, utilizing parallel Channel Feature Focusing Blocks (CFFB) and Spatial Feature Focusing Blocks (SFFB) to extract complementary features for fine textures and global structures in degraded regions. This dual-module design enables CMAMRNet to maintain precise mask guidance while achieving comprehensive feature representation for accurate mural restoration.
The main contributions of this work can be summarized as follows.

(1) We propose CMAMRNet, a novel mask-aware framework for mural restoration that maintains comprehensive mask guidance throughout the processing pipeline while effectively capturing multiscale contextual features through transformer architectures.

(2) We propose MAUDS that integrates mask information through dedicated channel operations, ensuring consistent mask sensitivity across resolution scales and preventing diminishing mask influence in deep networks.

(3) We introduce CFA operating at highest and lowest resolutions, employing parallel CFFB and SFFB modules to extract complementary features that capture both fine-grained textures and global structures in degraded regions.


\section{Methodology}
\subsection{Network Backbone Design}
As illustrated in~\cref{fig:CMAMRNet}, our network backbone adopts a U-shaped hierarchical architecture that systematically integrates Restormer~\cite{zamir2022restormer} blocks for multi-scale feature representation. In the encoder path, the features are progressively processed through three distinct resolution stages. Each stage employs Restormer encoder blocks that integrate Multi-Head Self-Attention (MHSA) and Feed-Forward Network (FFN) layers. The core operations within a Restormer encoder block can be formulated as:
\begin{equation}
\begin{aligned}
    \hat{x} &= x + \text{MHSA}(\text{LN}(x)),\\
    y &= \hat{x} + \text{FFN}(\text{LN}(\hat{x})),
\end{aligned}
\end{equation}
where $x$ denotes the input features, LN represents Layer Normalization, and $y$ is the output features. The MHSA operation can be further decomposed as:
\begin{equation}
\begin{aligned}
    \text{MHSA}(x) &= \text{Concat}(\text{head}_1,...,\text{head}_h)W^O, \\
    \text{head}_i &= \text{Attention}(xW_i^Q, xW_i^K, xW_i^V),
\end{aligned}
\end{equation}
where $W^Q$, $W^K$, $W^V$ are learnable projection matrices for query, key, and value respectively, and $W^O$ is the output projection matrix. In the decoder path, our network employs symmetric Restormer blocks to progressively reconstruct high-quality features through skip connections with the encoder.

A distinctive feature of our architecture is its comprehensive incorporation of mask information on multiple scales, ensuring effective degradation guidance throughout the restoration process. To maintain mask guidance, we propose the Mask-Aware Up/Down-Sampler (MAUDS) module at resolution transitions that ensures consistent mask sensitivity through dedicated channel operations. Furthermore, we introduce the Co-Feature Aggregator (CFA) module operating at both the highest and lowest resolutions, which extracts complementary features to capture fine-grained textures and global structures in degraded regions.
\subsection{Mask-Aware Up/Down-Sampler (MAUDS)}
\begin{figure}[!h]
    \centering
    \includegraphics[width=.65\linewidth]{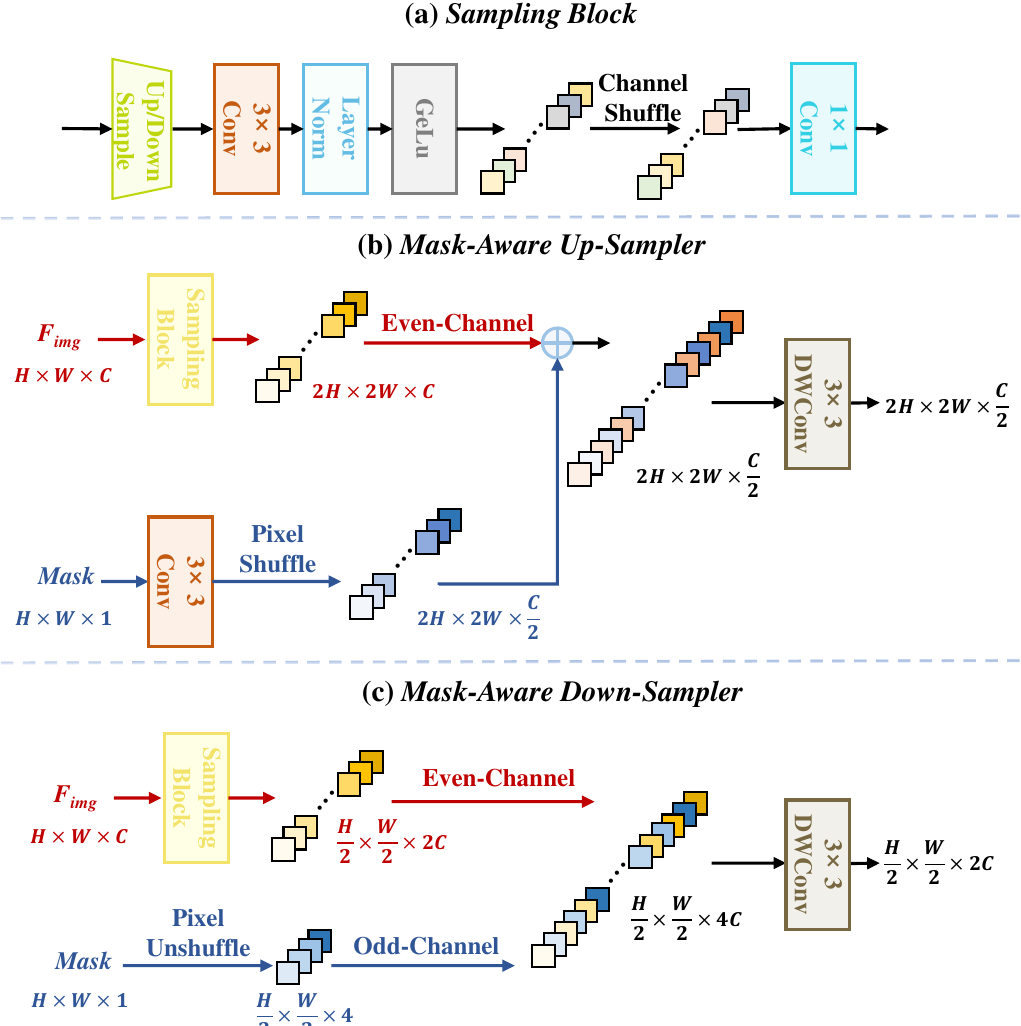}
    \caption{Overview of Mask-Aware Up/Down-Sampler (MAUDS): (a) Sampling Block, (b) Mask-Aware Down-Sampler (MADS), and (c) Mask-Aware Up-Sampler (MAUS) for mask-guided feature transitions across different resolution scales.}
    \label{fig:MAUDS}
\end{figure}
MAUDS consists of Mask-Aware Up-Sampler (MAUS) and Mask-Aware Down-Sampler (MADS) to maintain mask-guided feature transitions across different resolution scales. As shown in~\cref{fig:MAUDS} (a), the Sampling Block serves as a fundamental component for both samplers, where up/down sampling operations first adjust the spatial resolution, followed by channel shuffle operations to enhance information flow across channels. While the Sampling Block processes the mural image features ($F_{img}$), the mask undergoes a separate but parallel path through pixel shuffle/unshuffle operations. This parallel processing ensures that both the image features and the mask information are properly aligned and scaled, enabling effective mask-guided restoration at different resolution levels.

For the MAUS module, let $F_{img} \in \mathbb{R}^{H \times W \times C}$ denote the input feature maps and $M \in \mathbb{R}^{H \times W \times 1}$ denote the binary mask, where $C$ is the number of channels and $H, W$ are spatial dimensions. As illustrated in~\cref{fig:MAUDS} (b), the upsampling process can be formulated as:
\begin{equation}
\begin{aligned}
    F_{img}^{up}  &= \phi_{up}(F_{img}) \in \mathbb{R}^{2H \times 2W \times C}, \\
    M^{up}        &= \xi_{up}(\omega(M)) \in \mathbb{R}^{2H \times 2W \times C/2},
\end{aligned}
\end{equation}
where $\phi_{up}(\cdot)$ denotes the feature upsampling operation, $\omega(\cdot)$ represents the convolution operation, and $\xi_{up}(\cdot)$ is the pixel shuffle operation.

The channel selection and feature fusion process can be expressed as:
\begin{equation}
\begin{aligned}
    F_{img}^{select} &= \{F_{img,i}^{up} | i = 1,3,...,C\} \in \mathbb{R}^{2H \times 2W \times C/2}, \\
    F_{img}^{out}    &= \gamma(F_{img}^{select} + M^{up}) \in \mathbb{R}^{2H \times 2W \times C/2},
\end{aligned}
\end{equation}
where $\gamma(\cdot)$ represents the depth-wise convolution operation for feature refinement. This formulation demonstrates how mask information is integrated with the selected feature channels to guide the upsampling process while maintaining spatial consistency.

Similarly, the MADS module performs downsampling operations while maintaining mask awareness. As depicted in~\cref{fig:MAUDS} (c), the downsampling process can be formulated as:
\begin{equation}
\begin{aligned}
    F_{img}^{down}  &= \phi_{down}(F_{img}) \in \mathbb{R}^{H/2 \times W/2 \times 2C}, \\
    M^{down}        &= \xi_{down}(M) \in \mathbb{R}^{H/2 \times W/2 \times 4},
\end{aligned}
\end{equation}
where $\phi_{down}(\cdot)$ represents the feature downsampling operation through the Sampling Block, and $\xi_{down}(\cdot)$ denotes pixel unshuffle operation.

The feature interleaving and fusion process can be expressed as:
\begin{equation}
F_{inter} = \mathcal{I}(F_{img}^{down}, M^{down}) \in \mathbb{R}^{H/2 \times W/2 \times 4C},
\end{equation}
where the interleaving operation $\mathcal{I}(\cdot)$ is defined as:
\begin{equation}
\begin{aligned}
    F_{inter,2i-1} &= \{F_{img,i}^{down} | i = 1,2,...,2C\},\\
    F_{inter,2i} &= \{M_{j}^{down} | j = i \bmod 4, i = 1,2,...,2C\}.
\end{aligned}
\end{equation}

The final output is obtained through:
\begin{equation}
F_{img}^{out} = \gamma(F_{inter}) \in \mathbb{R}^{H/2 \times W/2 \times 2C},
\end{equation}
where $\gamma(\cdot)$ denotes the depth-wise convolution that processes the interleaved features to produce the final result. This design ensures effective integration of mask information during the downsampling process while maintaining feature coherence.

\begin{figure}[ht]
    \centering
    \includegraphics[width=.65\linewidth]{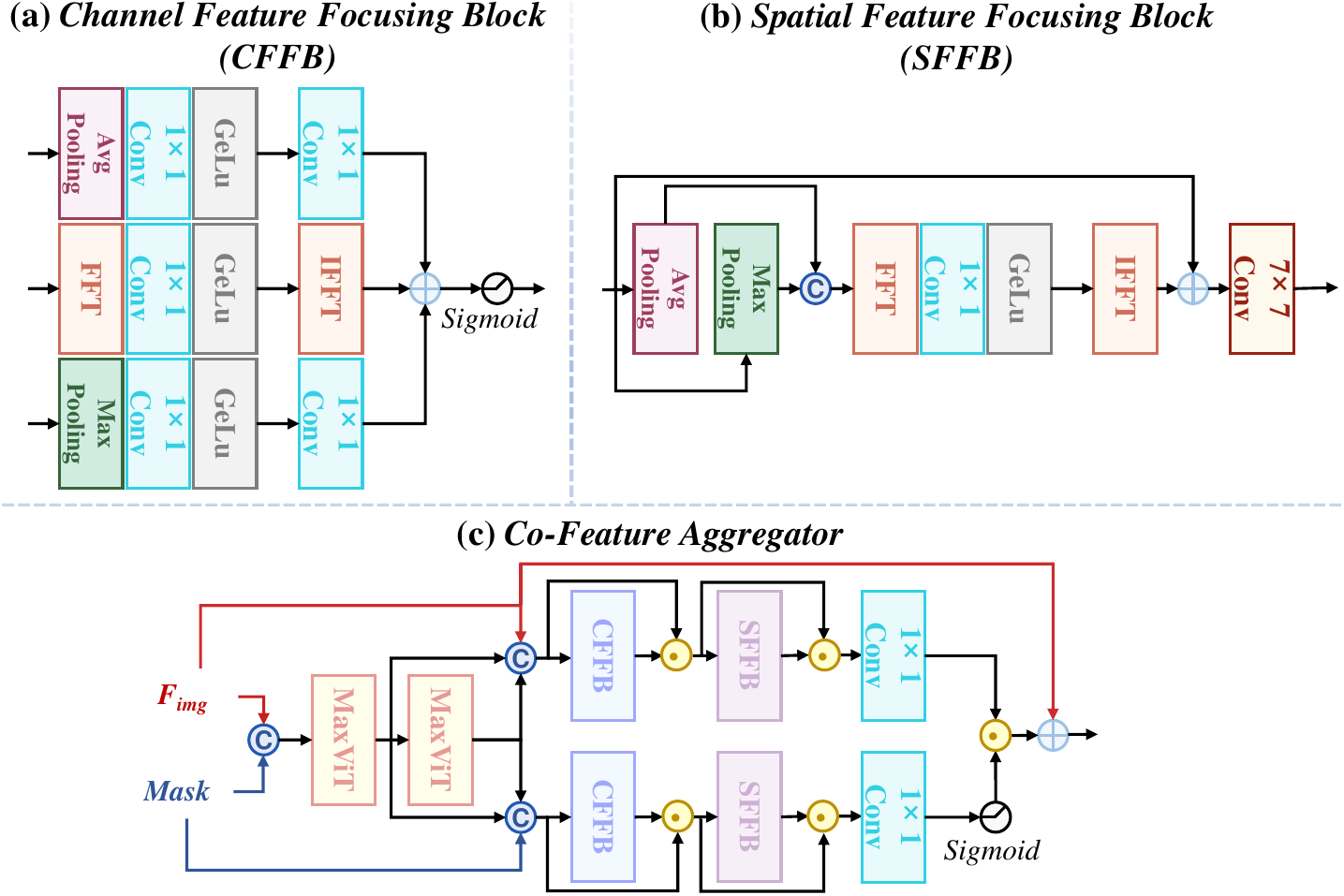}
    \caption{Architecture of: (a) the Channel Feature Focusing Block (CFFB), (b) the Spatial Feature Focusing Block (SFFB), and (c) the Co-Feature Aggregator (CFA) structure with integrated focusing blocks.}
    \label{fig:CFA}
\end{figure}
\subsection{Co-Feature Aggregator (CFA)}
The Co-Feature Aggregator (CFA), shown in~\cref{fig:CFA}, employs parallel Channel Feature Focusing Blocks (CFFB) and Spatial Feature Focusing Blocks (SFFB) to extract complementary features. It combines fine-grained textures and global structures while preserving original feature information through adaptive feature modulation and residual connections. For input feature maps $F_{img} \in \mathbb{R}^{H \times W \times C}$ and mask $M \in \mathbb{R}^{H \times W \times 1}$, the feature extraction process can be formulated as:
\begin{equation}
\begin{aligned}
    F_1 &= \text{MaxViT}(\text{Concat}[F_{img}, M]), \\
    F_2 &= \text{MaxViT}(F_1),
\end{aligned}
\end{equation}
where MaxViT~\cite{tu2022maxvit} blocks process the concatenated features and intermediate features respectively.

The CFA performs multi-level feature aggregation through feature focusing blocks:
\begin{equation}
\begin{aligned}
    F_{cat\_rgb} &= \text{Concat}[F_1, F_2, F_{img}], \\
    F_{cat\_mask} &= \text{Concat}[F_1, F_2, M].
\end{aligned}
\end{equation}

Each concatenated feature undergoes sequential channel and spatial attention refinement:
\begin{equation}
\begin{aligned}
F_{mid} &= F_{in} \odot \text{CFFB}(F_{in}), \\
F_{out} &= F_{mid} \odot \text{SFFB}(F_{mid}),
\end{aligned}
\end{equation}
where $\odot$ denotes element-wise multiplication. The features are then followed by dimension reduction through 1×1 convolution to obtain $F_{rgb}$ and $F_{mask}$ respectively. As illustrated in~\cref{fig:CFA} (a) and (b), in both CFFB and SFFB, Fast Fourier Transform (FFT) operations help capture frequency patterns of image textures and mask degradation distributions, enabling comprehensive feature extraction in both spatial and frequency domains. 

The final CFA output is obtained via adaptive feature modulation:
\begin{equation}
    F_{cfa} = F_{rgb} \odot \sigma(F_{mask}) + F_{img},
\end{equation}
where $\sigma(\cdot)$ is the sigmoid activation function, and the residual connection preserves the original feature information.
\subsection{Loss Function}
Our restoration framework combines two complementary objectives: Mean Square Error (MSE) for pixel-level accuracy and Structural Similarity Index Measure (SSIM)~\cite{wang2004image} for structural fidelity. The MSE loss performs pixel-wise comparison between the restored and ground truth images:
\begin{equation}
\mathcal{L}_{MSE} = \frac{1}{N}\sum_{i=1}^{N}(x_i - \hat{x}_i)^2,
\end{equation}
where $x_i$ and $\hat{x}_i$ denote the pixel values of ground truth and restored images respectively, and $N$ is the total number of pixels.

To better preserve structural information and perceptual quality, we incorporate the SSIM loss:
\begin{equation}
\mathcal{L}_{SSIM} = 1 - \frac{(2\mu_x\mu_{\hat{x}} + C_1)(2\sigma_{x\hat{x}} + C_2)}{(\mu_x^2 + \mu_{\hat{x}}^2 + C_1)(\sigma_x^2 + \sigma_{\hat{x}}^2 + C_2)},
\end{equation}
where $\mu_x$, $\mu_{\hat{x}}$ are means, $\sigma_x^2$, $\sigma_{\hat{x}}^2$ are variances, $\sigma_{x\hat{x}}$ is covariance, and $C_1$, $C_2$ are stability constants.

The total loss function combines both terms with a weighting factor $\lambda$:
\begin{equation}
\mathcal{L}_{total} = \mathcal{L}_{MSE} + \lambda\mathcal{L}_{SSIM},
\end{equation}
where we empirically set $\lambda=0.4$ to balance pixel-accurate restoration and structural consistency.
\begin{table}[th]
    \centering
    \caption{Quantitative results of comparative experiments with state-of-the-art methods. The best results are shown in \textbf{bold} and the second best results are \underline{underlined}.}
    \begin{adjustbox}{max width=\linewidth}
    \begin{tabular}{c|cccc|cccc}
    \toprule
    \multirow{2}{*}{Method}&
    \multicolumn{4}{c|}{MuralDH~\cite{xu2024comprehensive}}&
    \multicolumn{4}{c}{Dunhuang~\cite{yu2019dunhuang}}\\
    \cmidrule(l){2-9}
    & PSNR $\uparrow$ & SSIM $\uparrow$ & MAE $\downarrow$ & LPIPS $\downarrow$
    & PSNR $\uparrow$ & SSIM $\uparrow$ & MAE $\downarrow$ & LPIPS $\downarrow$\\
    \midrule
    RN~\cite{yu2020region} & 14.3577 & 0.2712 & 42.0932 & 0.9877 & 17.9298 & 0.5401 & 23.9331 & 0.7688 \\
    DeepFillV2~\cite{yu2019free}  & 14.4970 & 0.2734 & 41.5719 & 1.0131 & 16.2900 & 0.5002 & 30.8472 & 0.8756 \\
    HFill~\cite{yi2020contextual} & 14.5221 & 0.2741 & 41.5619 & 1.0232 & 17.0136 & 0.5058 & 26.3423 & 0.9000 \\
    CoordFill~\cite{liu2023coordfill} & 23.6762 & 0.5706 & 12.2239 & 0.3160 & 20.4858 & 0.4134 & 16.1557 & 0.5367 \\
    SyFormer~\cite{wu2024syformer} & 24.5005 & 0.6328 & 11.3667 & 0.6102 & 26.0647 & 0.7156 & 8.8938 & 0.4659 \\
    LaMa~\cite{suvorov2022resolution} & 28.4603 & 0.9195 & 6.6463 & 0.0861 & 29.5888 & 0.8346 & 5.8470 & 0.1067 \\
    EdgeConnect~\cite{nazeri2019edgeconnect} & 29.9298 & 0.9427 & 5.4444 & 0.0788 & 30.7807 & 0.8519 & 4.9383 & 0.1045 \\
    T-Former~\cite{deng2022t} & 30.9527 & 0.9564 & 4.5003 & 0.0729 & 31.5000 & 0.8608 & 4.6571 & 0.0854 \\
    DWTNet~\cite{shamsolmoali2024distance} & 31.0443 & 0.9565 & 4.4492 & 0.0726 & 31.5097 & 0.8609 & 4.6609 & 0.0855 \\
    Misf~\cite{li2022misf} & 34.1878 & 0.9635 & 1.8124 & 0.0555 & 32.1570 & \underline{0.8701} & 4.4660 & \underline{0.0691} \\
    DFNet~\cite{hong2019deep} & 34.6129 & 0.9641 & 1.3644 & 0.0577 & 31.0897 & 0.8549 & 4.7918 & 0.0936 \\
    HINT~\cite{chen2024hint} & \underline{35.4833} & \underline{0.9663} & \underline{1.2337} & \underline{0.0542} & 32.2421 & 0.8695 & \underline{4.3776} & 0.0706 \\
    \midrule
    \rowcolor[gray]{0.9} \textbf{Ours} &\textbf{36.2565} & \textbf{0.9683} & \textbf{1.1454}& \textbf{0.0497}& \textbf{33.0738}& \textbf{0.8944} & \textbf{3.9708}& \textbf{0.0598} \\
    \bottomrule
    \end{tabular}
    \end{adjustbox}
    \label{table:Compare}
\end{table}
\section{Experiment}
\begin{figure}[t]
    \centering
    \begin{minipage}[t]{1.\linewidth}
        \begin{minipage}[t]{0.105\linewidth}
            \centering
            \includegraphics[width=\linewidth]{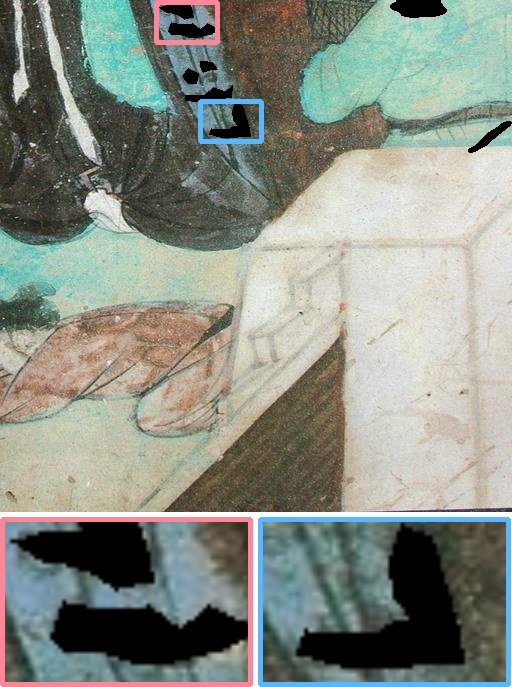}
        \end{minipage}
        \hfill
        \begin{minipage}[t]{0.105\linewidth}
            \centering
            \includegraphics[width=\linewidth]{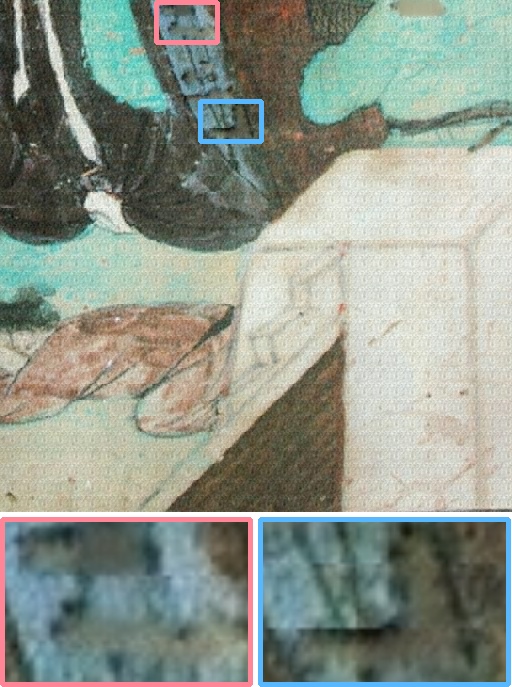}
        \end{minipage}
        \hfill
        \begin{minipage}[t]{0.105\linewidth}
            \centering
            \includegraphics[width=\linewidth]{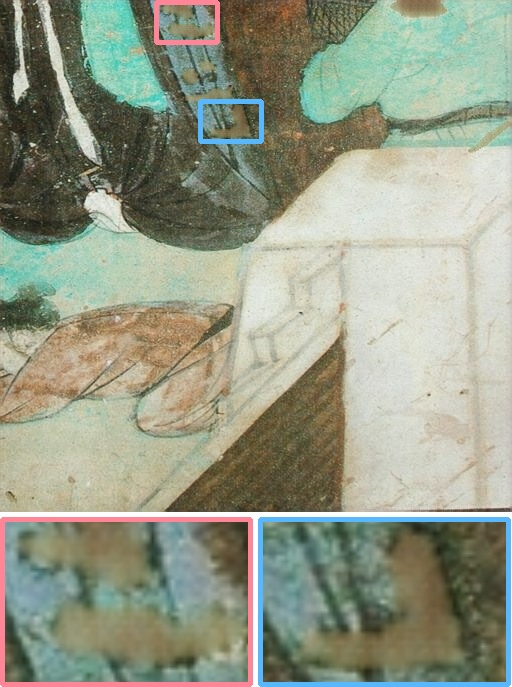}
        \end{minipage}
        \hfill
        \begin{minipage}[t]{0.105\linewidth}
            \centering
            \includegraphics[width=\linewidth]{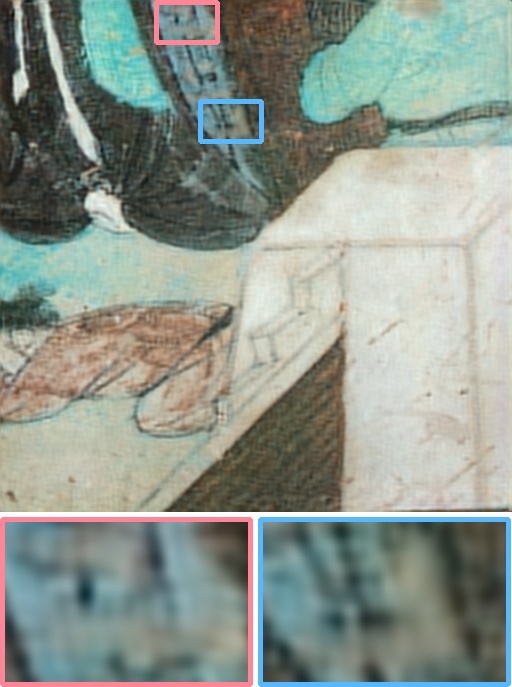}
        \end{minipage}
        \hfill
        \begin{minipage}[t]{0.105\linewidth}
            \centering
            \includegraphics[width=\linewidth]{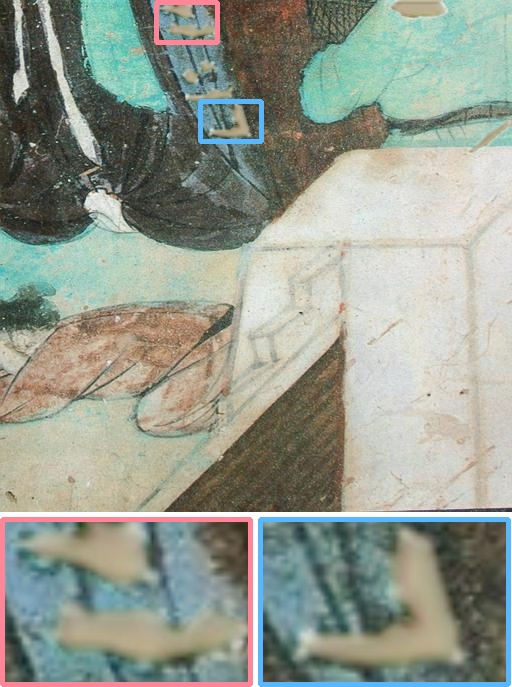}
        \end{minipage} 
        \hfill
        \begin{minipage}[t]{0.105\linewidth}
            \centering
            \includegraphics[width=\linewidth]{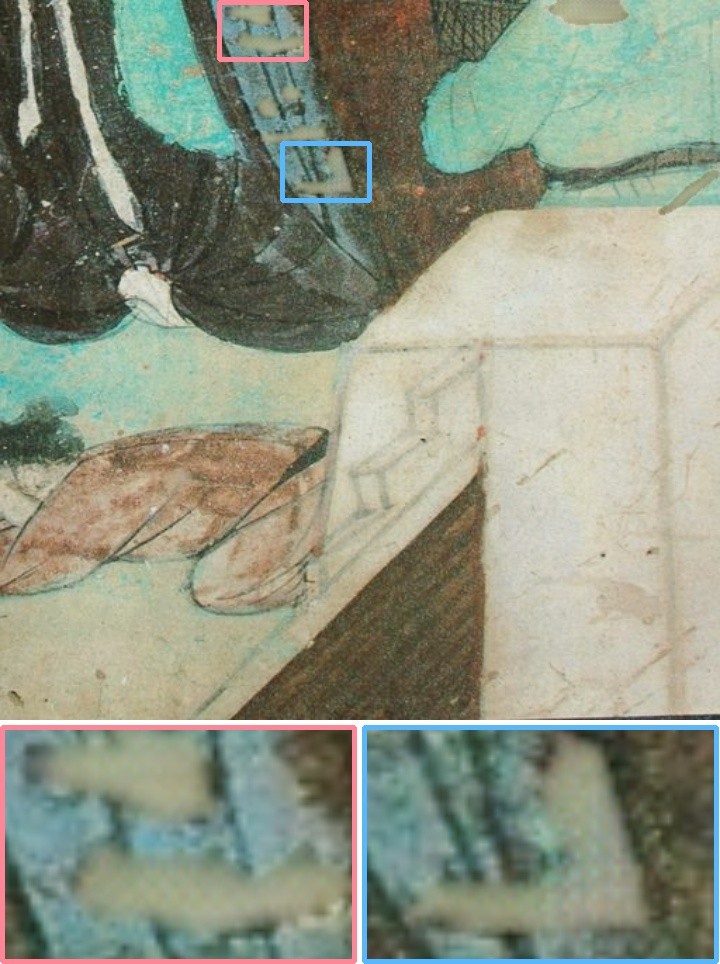}
        \end{minipage}  
        \hfill
        \begin{minipage}[t]{0.105\linewidth}
            \centering
            \includegraphics[width=\linewidth]{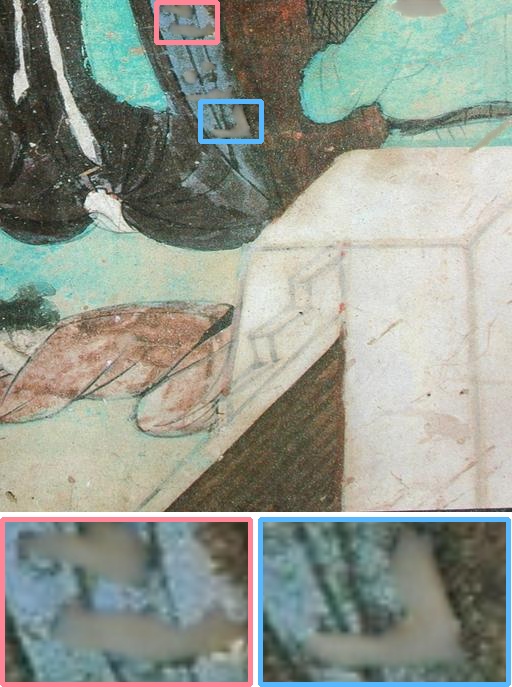}
        \end{minipage}  
        \hfill
        \begin{minipage}[t]{0.105\linewidth}
            \centering
            \includegraphics[width=\linewidth]{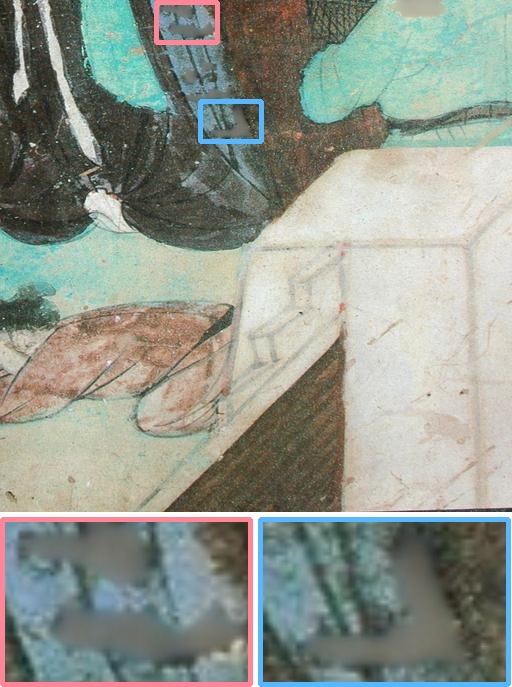}
        \end{minipage}
        \hfill
        \begin{minipage}[t]{0.105\linewidth}
            \centering
            \includegraphics[width=\linewidth]{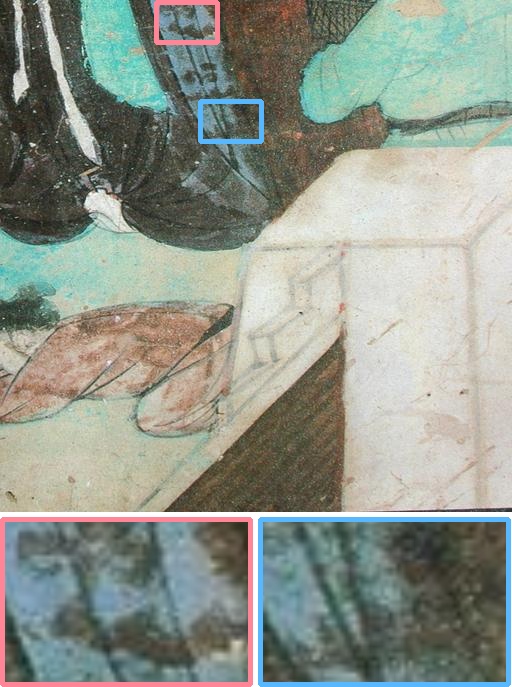}
        \end{minipage}
    \end{minipage}
    \begin{minipage}[t]{1.\linewidth}
        \begin{minipage}[t]{0.105\linewidth}
            \centering
            \includegraphics[width=\linewidth]{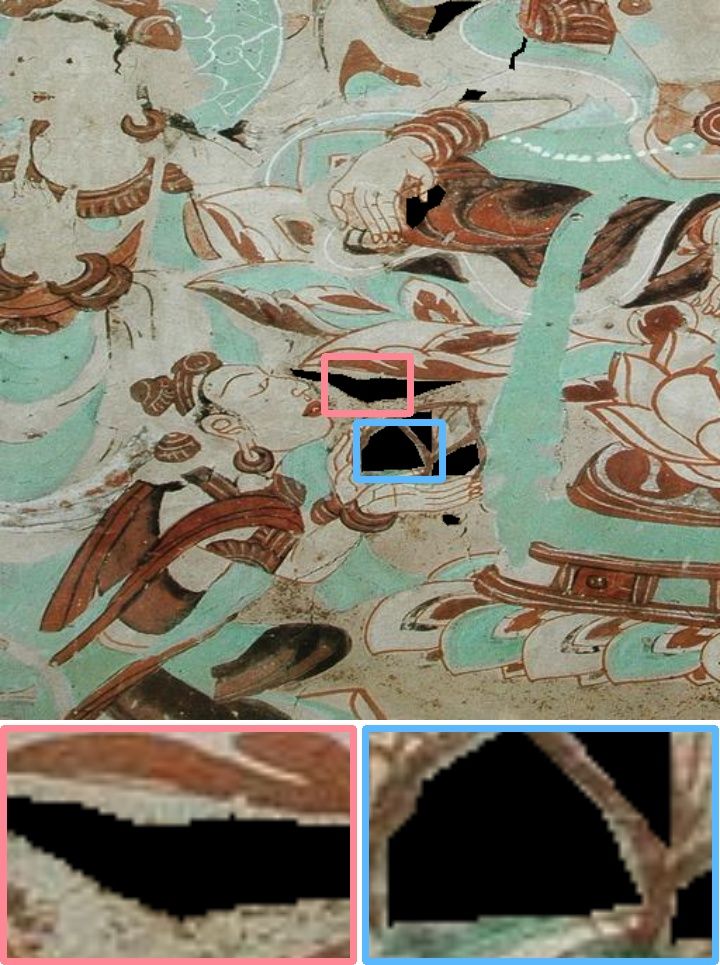}
        \end{minipage}
        \hfill
        \begin{minipage}[t]{0.105\linewidth}
            \centering
            \includegraphics[width=\linewidth]{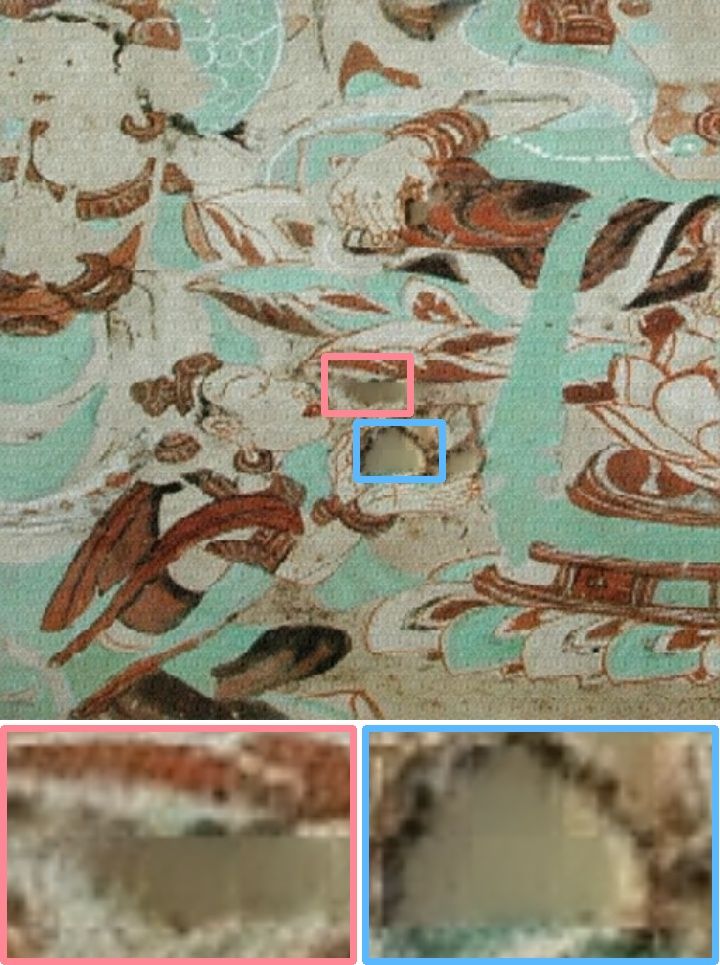}
        \end{minipage}
        \hfill
        \begin{minipage}[t]{0.105\linewidth}
            \centering
            \includegraphics[width=\linewidth]{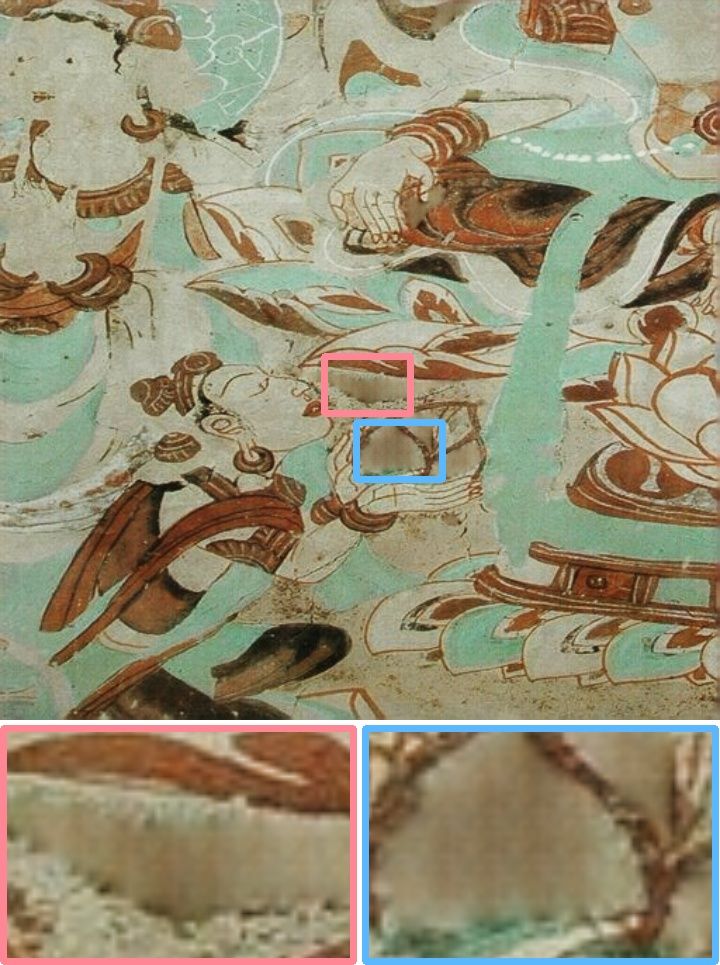}
        \end{minipage}
        \hfill
        \begin{minipage}[t]{0.105\linewidth}
            \centering
            \includegraphics[width=\linewidth]{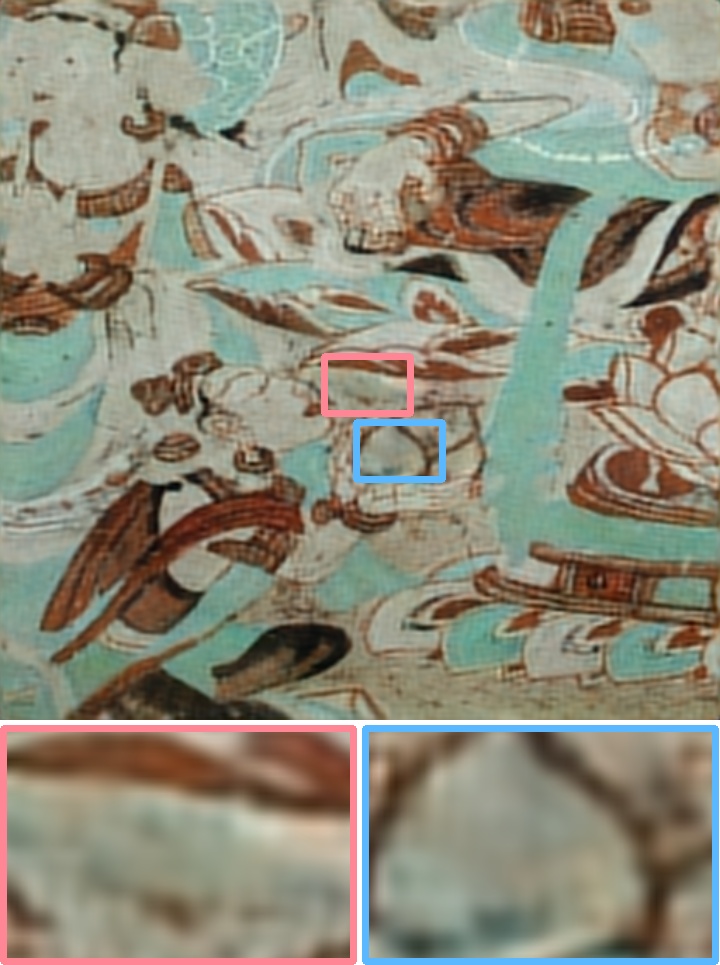}
        \end{minipage}
        \hfill
        \begin{minipage}[t]{0.105\linewidth}
            \centering
            \includegraphics[width=\linewidth]{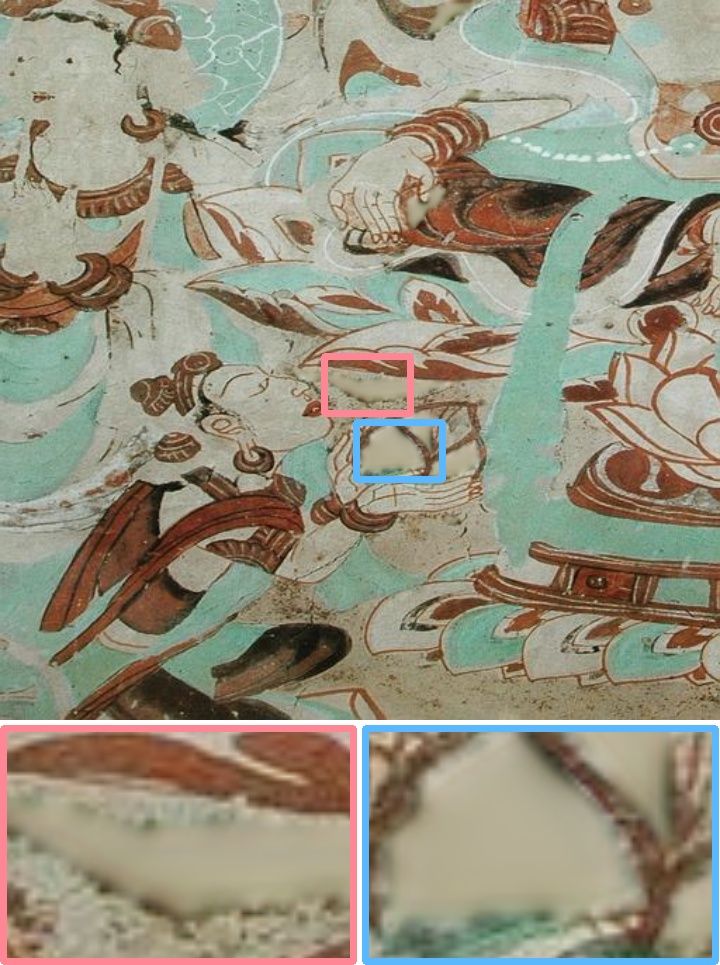}
        \end{minipage} 
        \hfill
        \begin{minipage}[t]{0.105\linewidth}
            \centering
            \includegraphics[width=\linewidth]{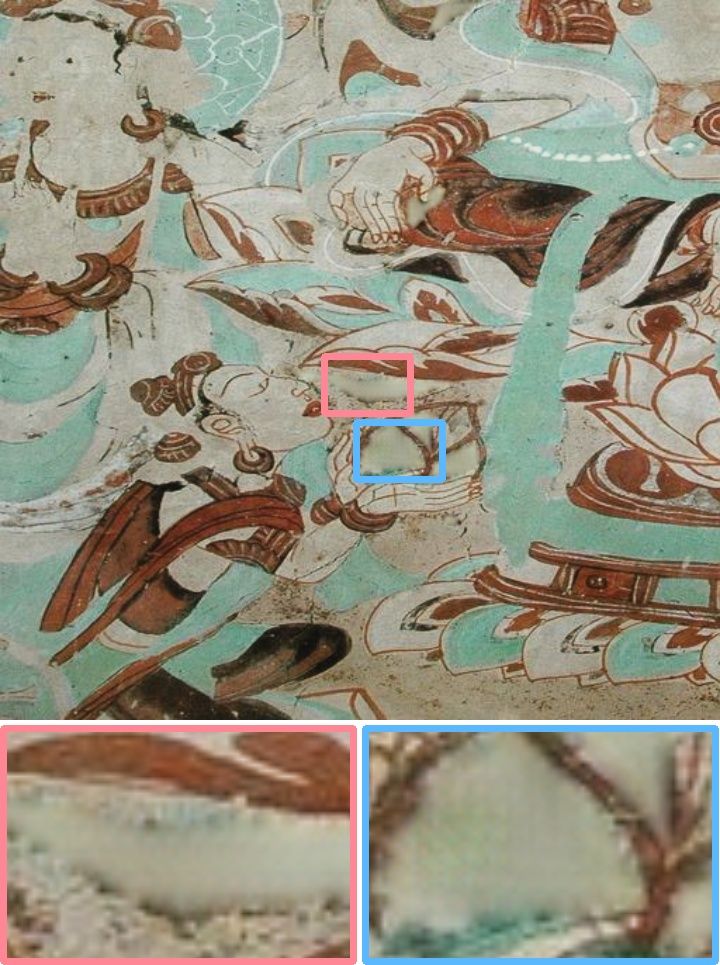}
        \end{minipage}  
        \hfill
        \begin{minipage}[t]{0.105\linewidth}
            \centering
            \includegraphics[width=\linewidth]{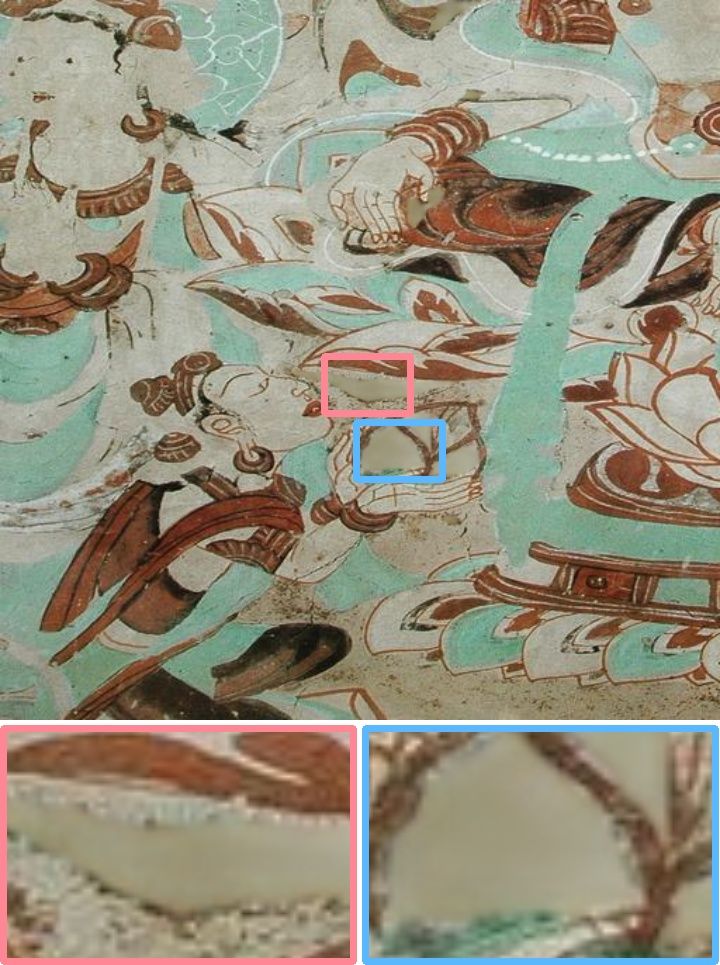}
        \end{minipage}  
        \hfill
        \begin{minipage}[t]{0.105\linewidth}
            \centering
            \includegraphics[width=\linewidth]{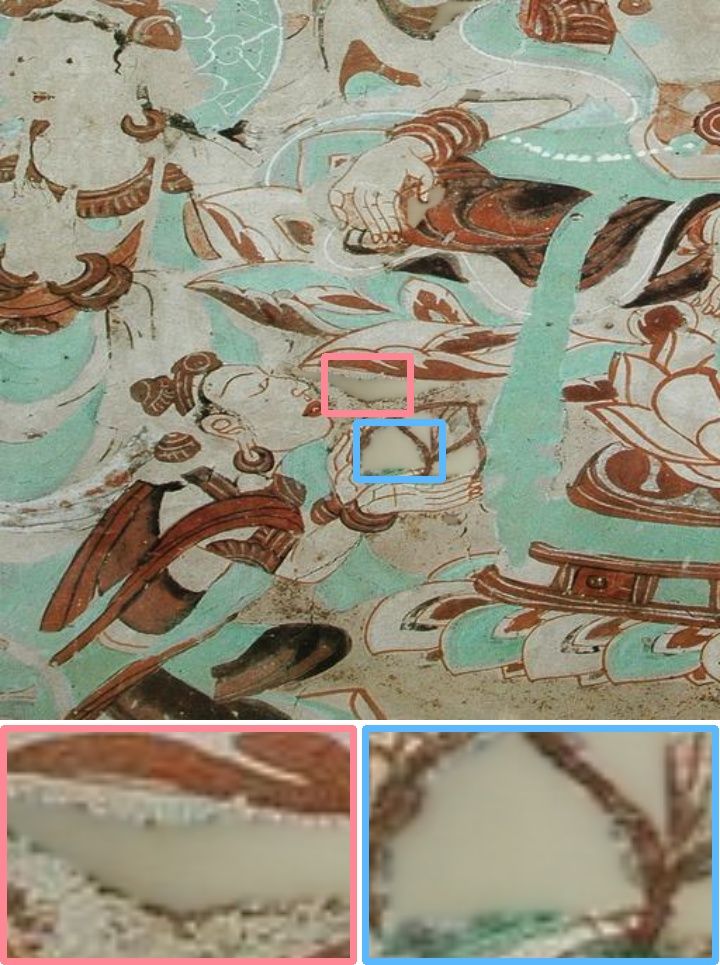}
        \end{minipage}
        \hfill
        \begin{minipage}[t]{0.105\linewidth}
            \centering
            \includegraphics[width=\linewidth]{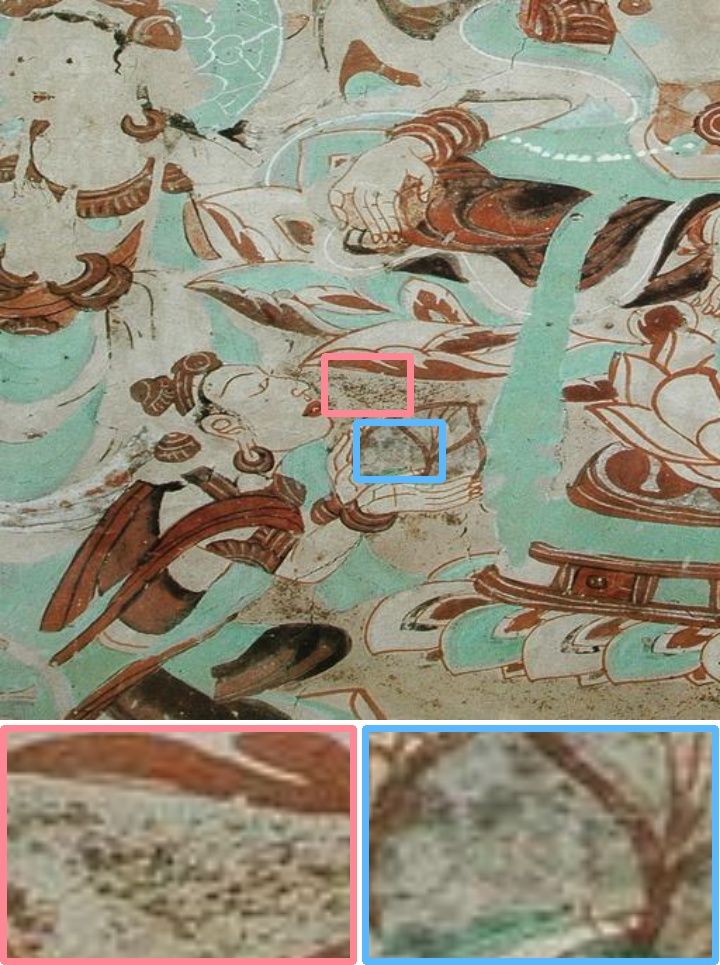}
        \end{minipage}
    \end{minipage}
    \begin{minipage}[t]{1.\linewidth}
        \begin{minipage}[t]{0.105\linewidth}
            \centering
            \includegraphics[width=\linewidth]{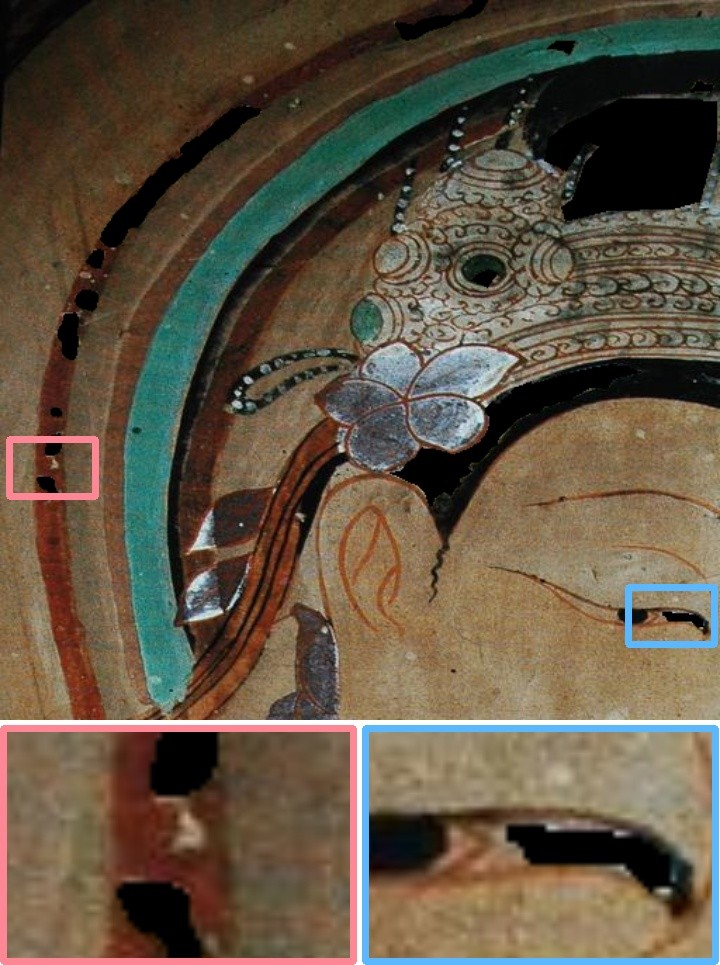}
        \end{minipage}
        \hfill
        \begin{minipage}[t]{0.105\linewidth}
            \centering
            \includegraphics[width=\linewidth]{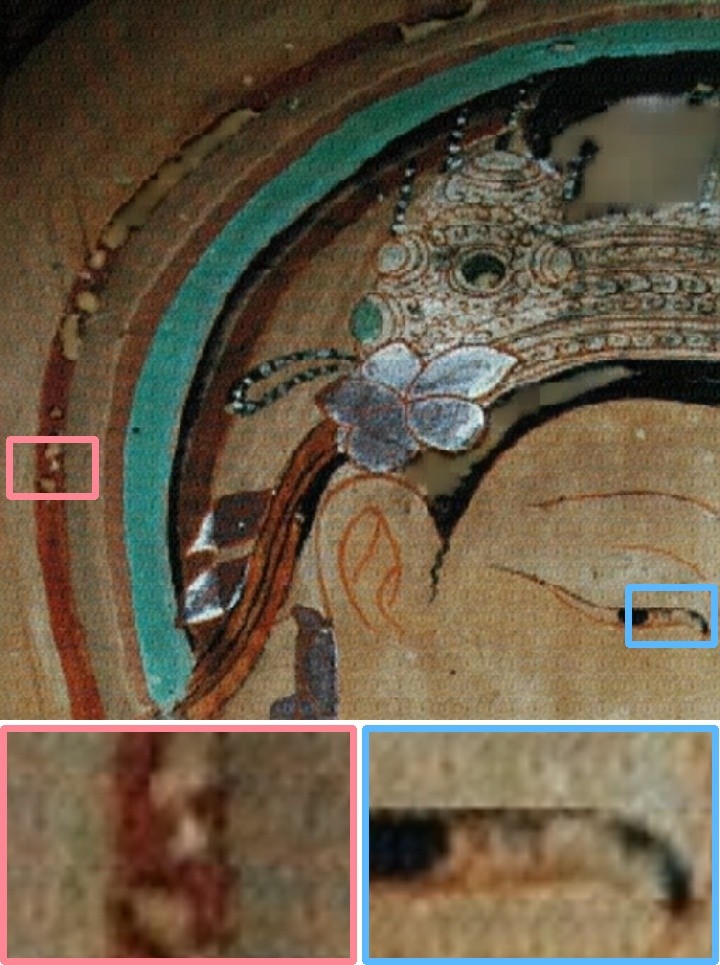}
        \end{minipage}
        \hfill
        \begin{minipage}[t]{0.105\linewidth}
            \centering
            \includegraphics[width=\linewidth]{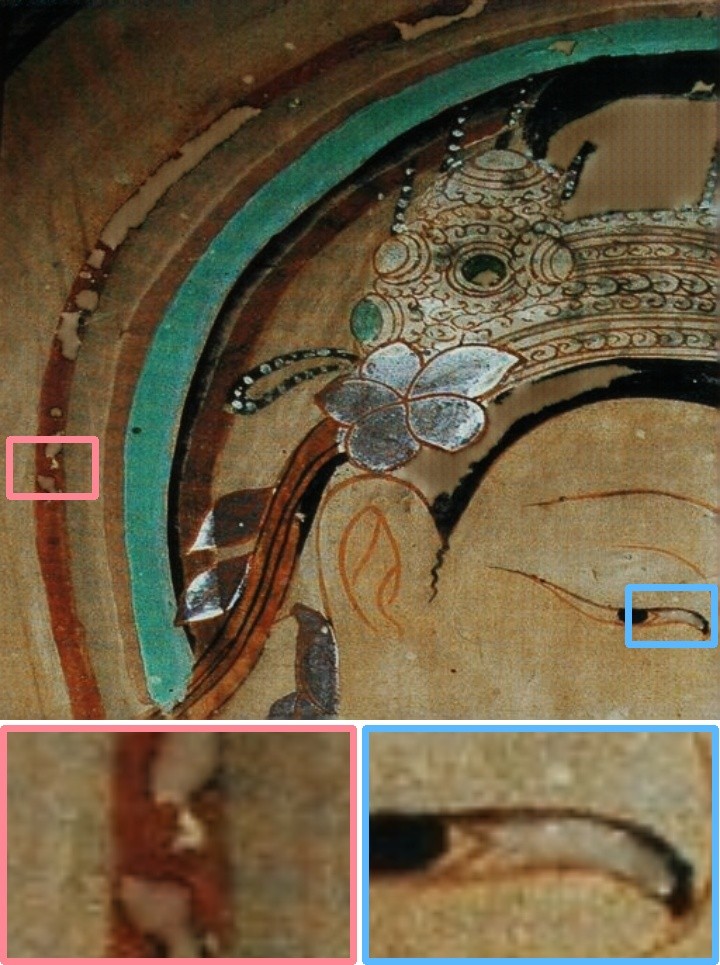}
        \end{minipage}
        \hfill
        \begin{minipage}[t]{0.105\linewidth}
            \centering
            \includegraphics[width=\linewidth]{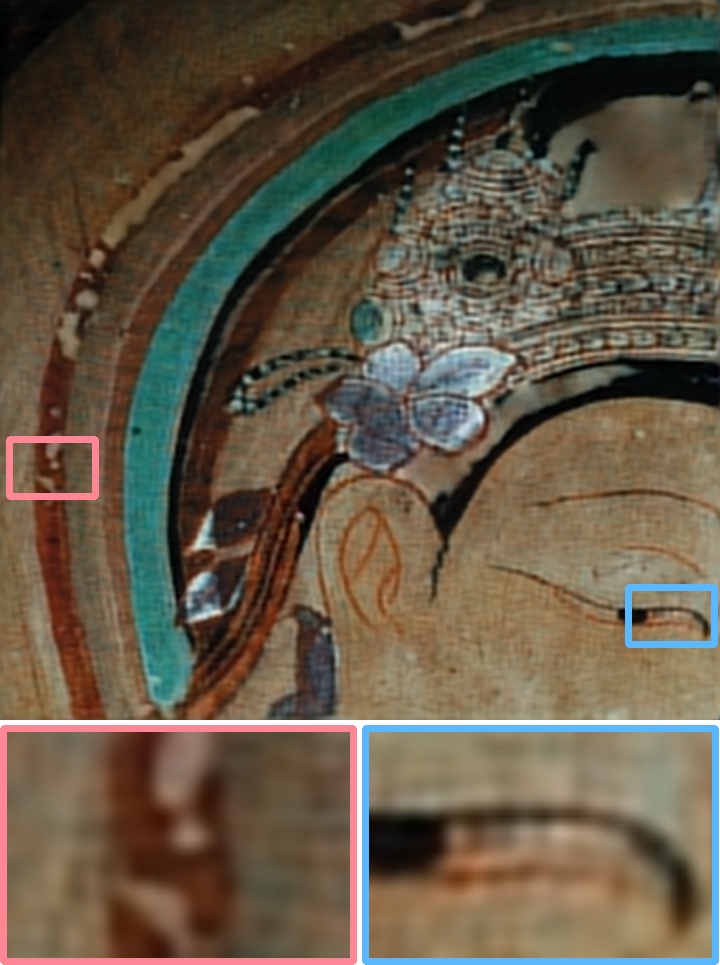}
        \end{minipage}
        \hfill
        \begin{minipage}[t]{0.105\linewidth}
            \centering
            \includegraphics[width=\linewidth]{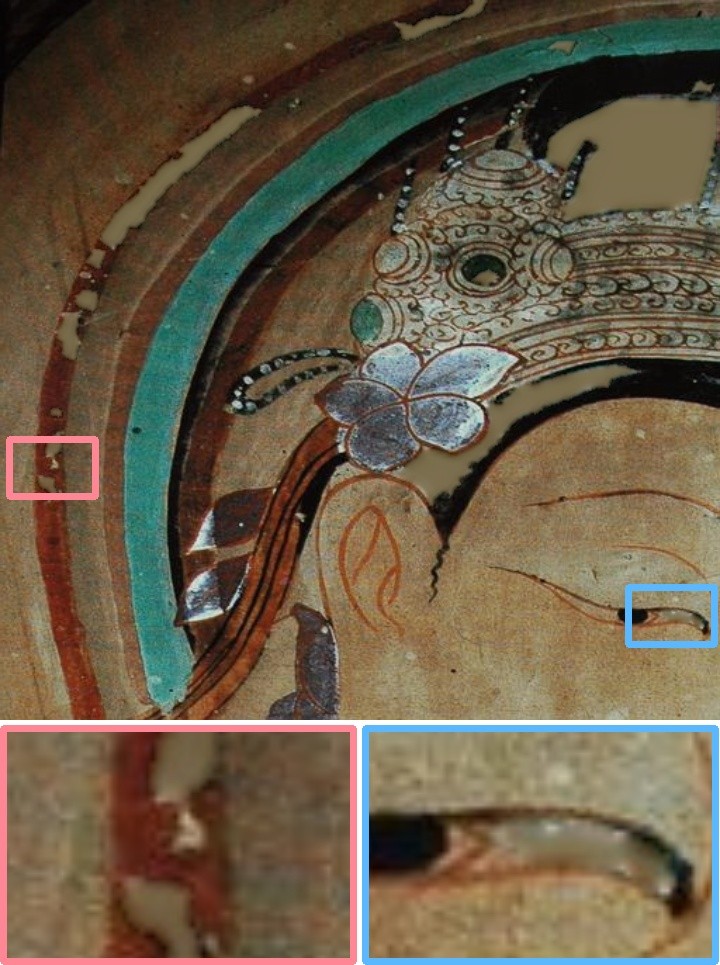}
        \end{minipage} 
        \hfill
        \begin{minipage}[t]{0.105\linewidth}
            \centering
            \includegraphics[width=\linewidth]{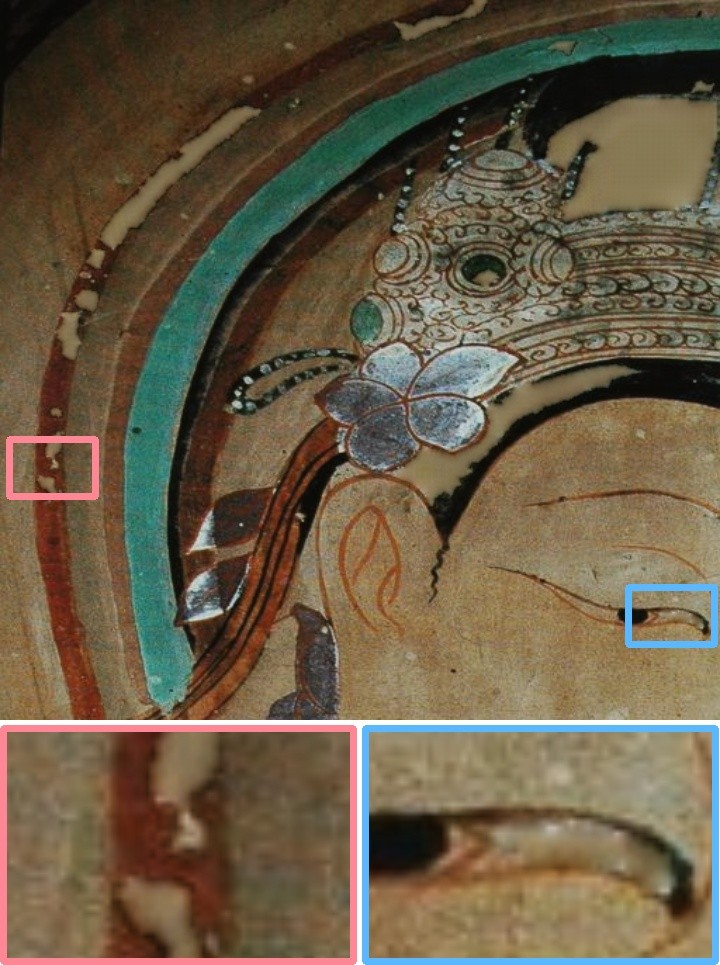}
        \end{minipage}  
        \hfill
        \begin{minipage}[t]{0.105\linewidth}
            \centering
            \includegraphics[width=\linewidth]{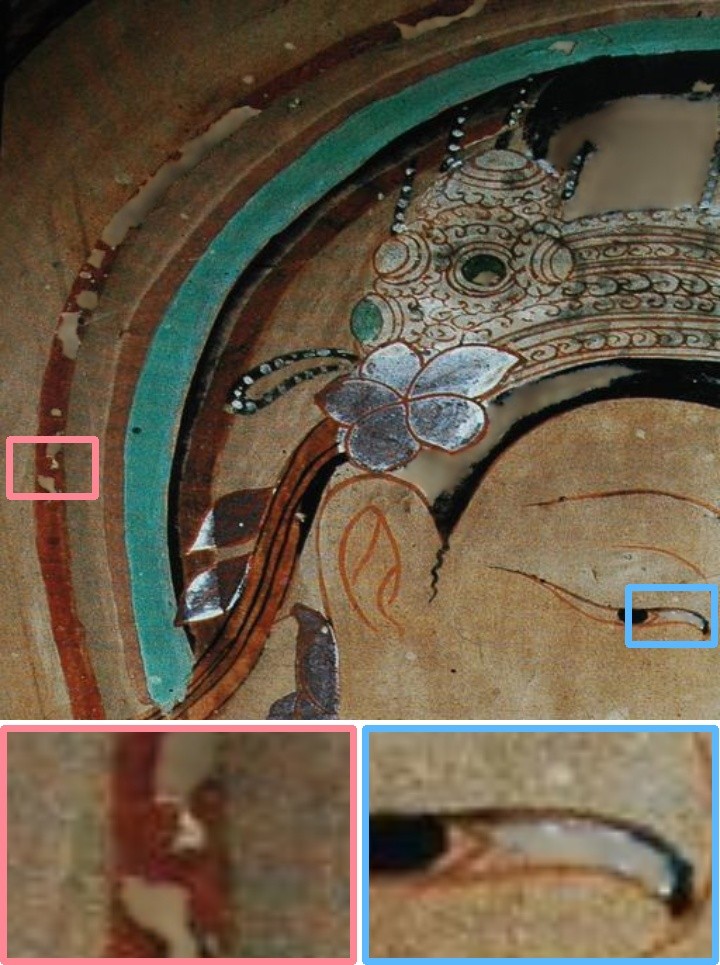}
        \end{minipage}  
        \hfill
        \begin{minipage}[t]{0.105\linewidth}
            \centering
            \includegraphics[width=\linewidth]{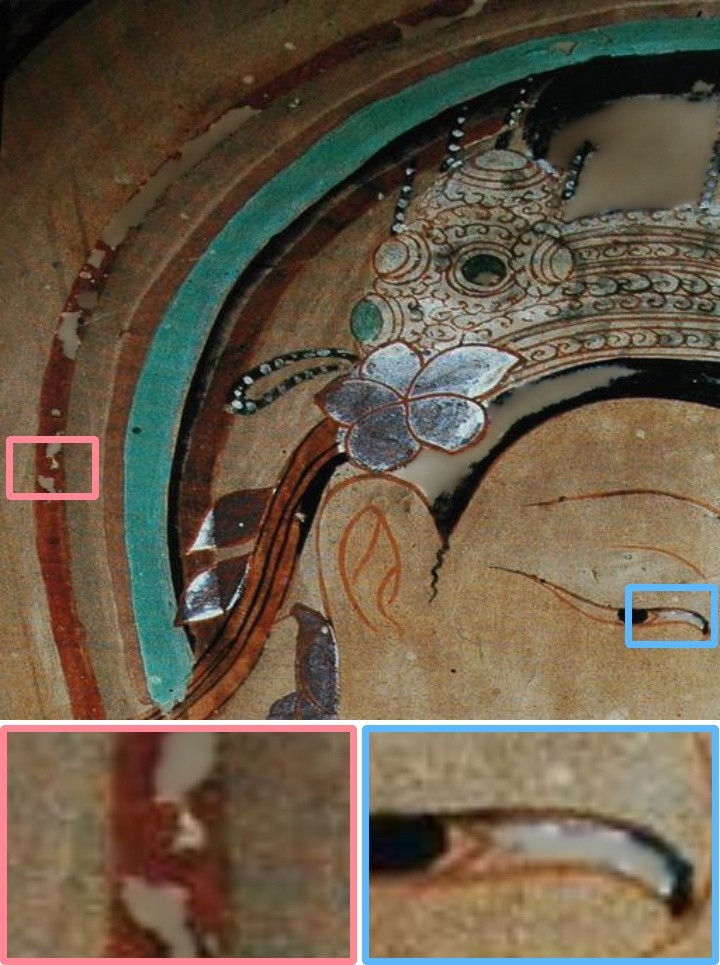}
        \end{minipage}
        \hfill
        \begin{minipage}[t]{0.105\linewidth}
            \centering
            \includegraphics[width=\linewidth]{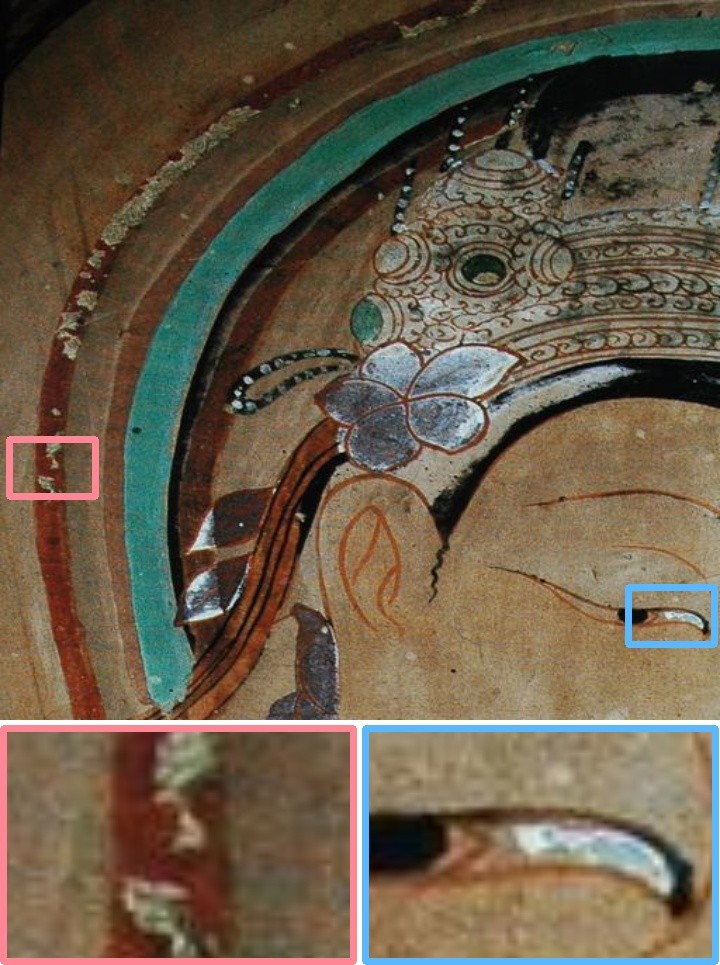}
        \end{minipage}
    \end{minipage}
    \begin{minipage}[t]{1.0\linewidth}
        \begin{minipage}[t]{0.105\linewidth}
            \centering
            \includegraphics[width=\linewidth]{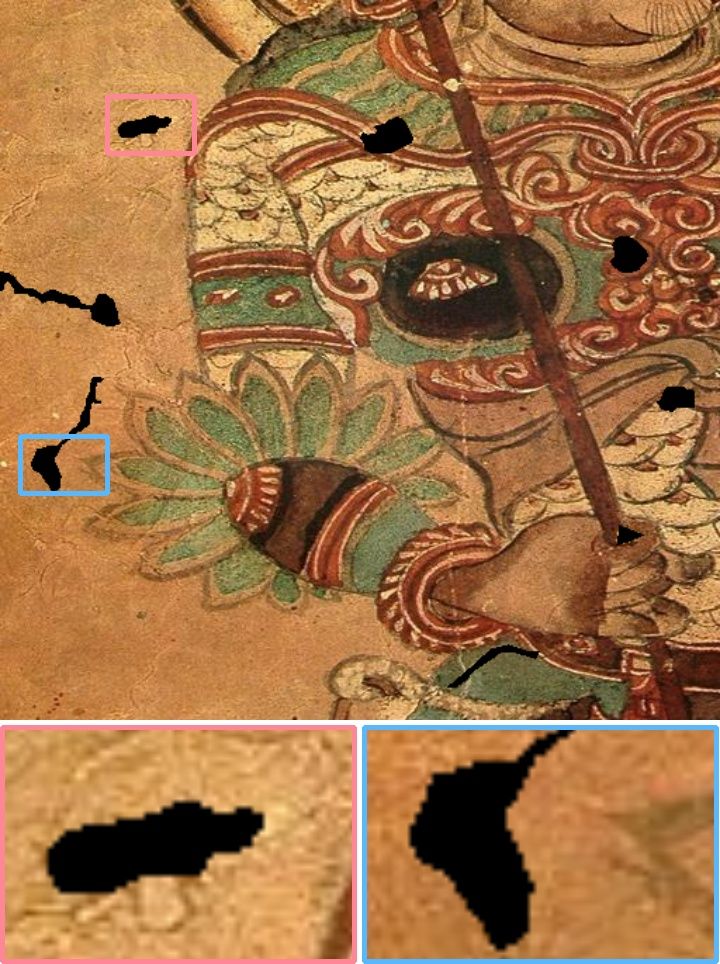}
            \centerline{\scriptsize (a) Input}
        \end{minipage}
        \hfill
        \begin{minipage}[t]{0.105\linewidth}
            \centering
            \includegraphics[width=\linewidth]{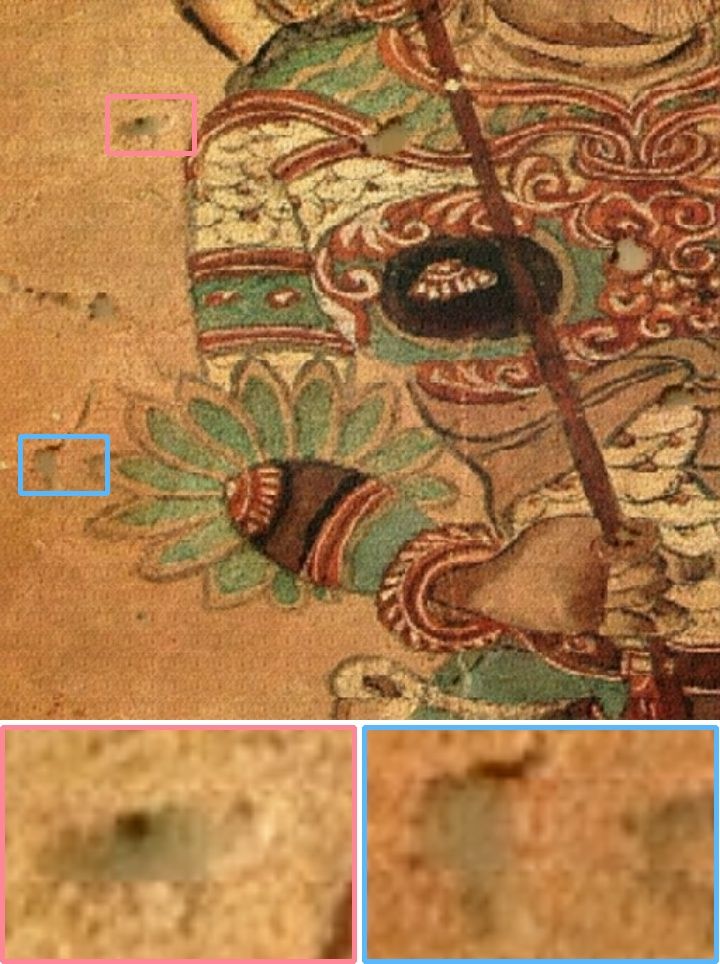}
            \centerline{\scriptsize (b) CoordFill}
        \end{minipage}  
        \hfill
        \begin{minipage}[t]{0.105\linewidth}
            \centering
            \includegraphics[width=\linewidth]{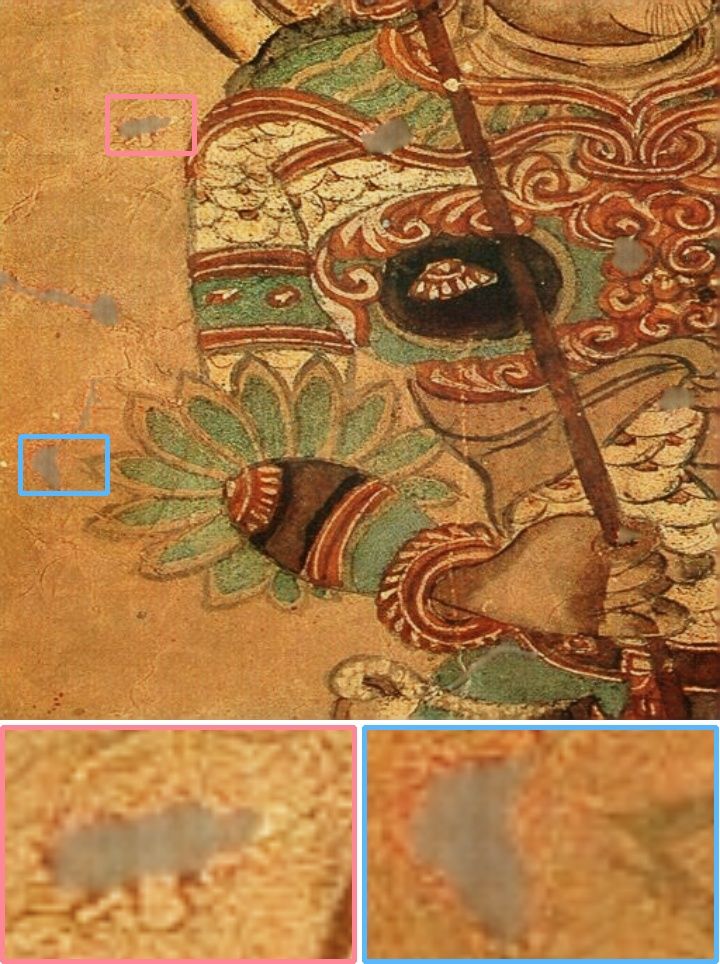}
            \centerline{\scriptsize (c) LaMa}
        \end{minipage}   
        \hfill
        \begin{minipage}[t]{0.105\linewidth}
            \centering
            \includegraphics[width=\linewidth]{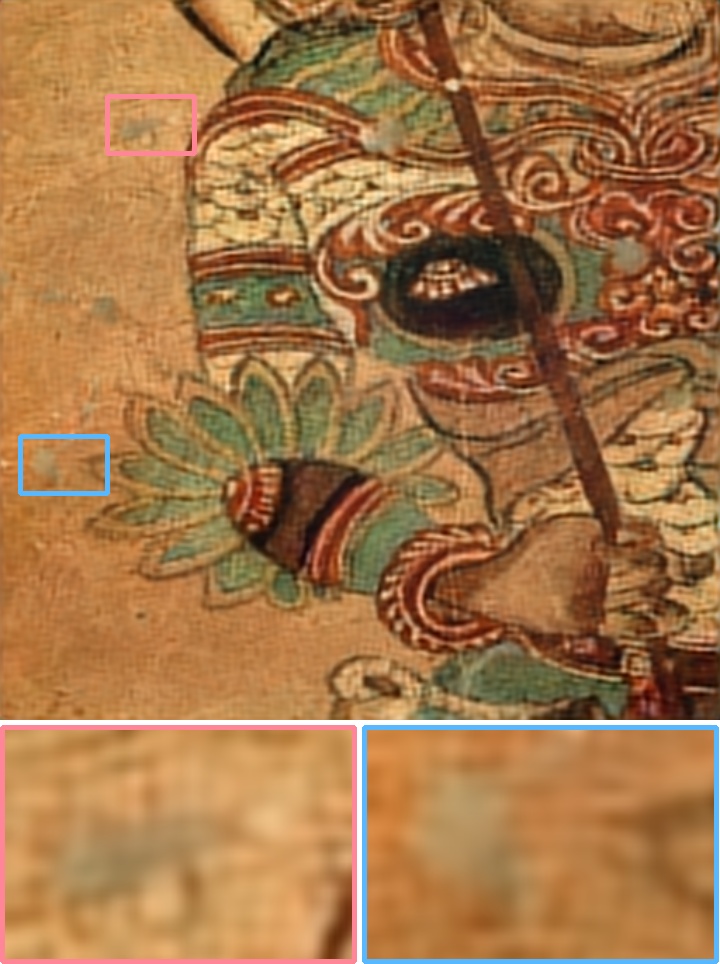}
            \centerline{\scriptsize (d) SyFormer}
        \end{minipage}
        \hfill
        \begin{minipage}[t]{0.105\linewidth}
            \centering
            \includegraphics[width=\linewidth]{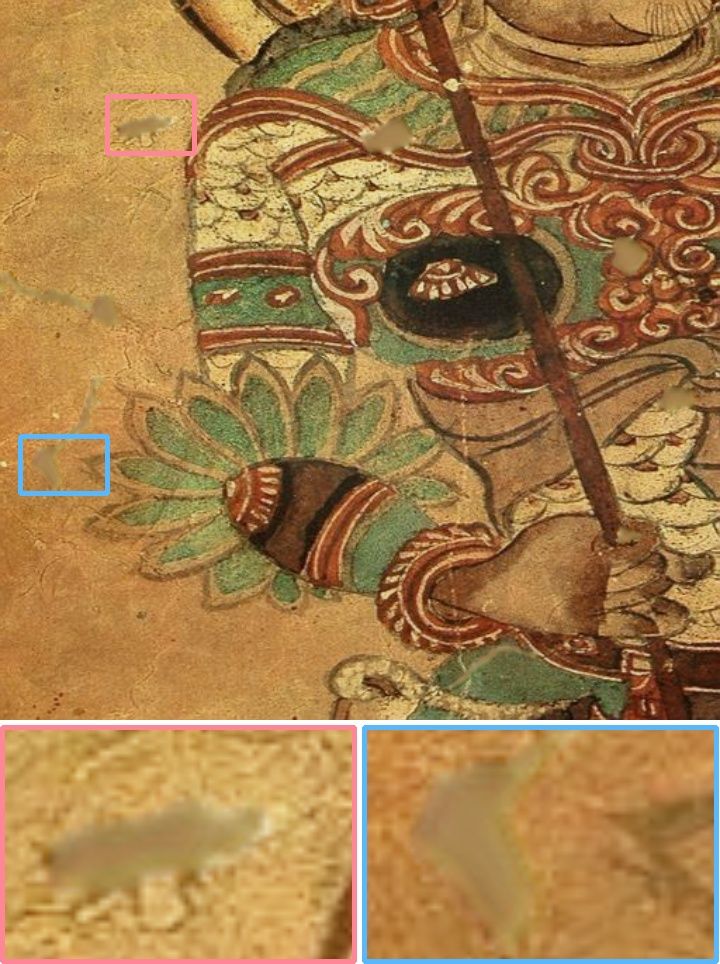}
            \centerline{\scriptsize (e) TFormer}
        \end{minipage}
        \hfill
        \begin{minipage}[t]{0.105\linewidth}
            \centering
            \includegraphics[width=\linewidth]{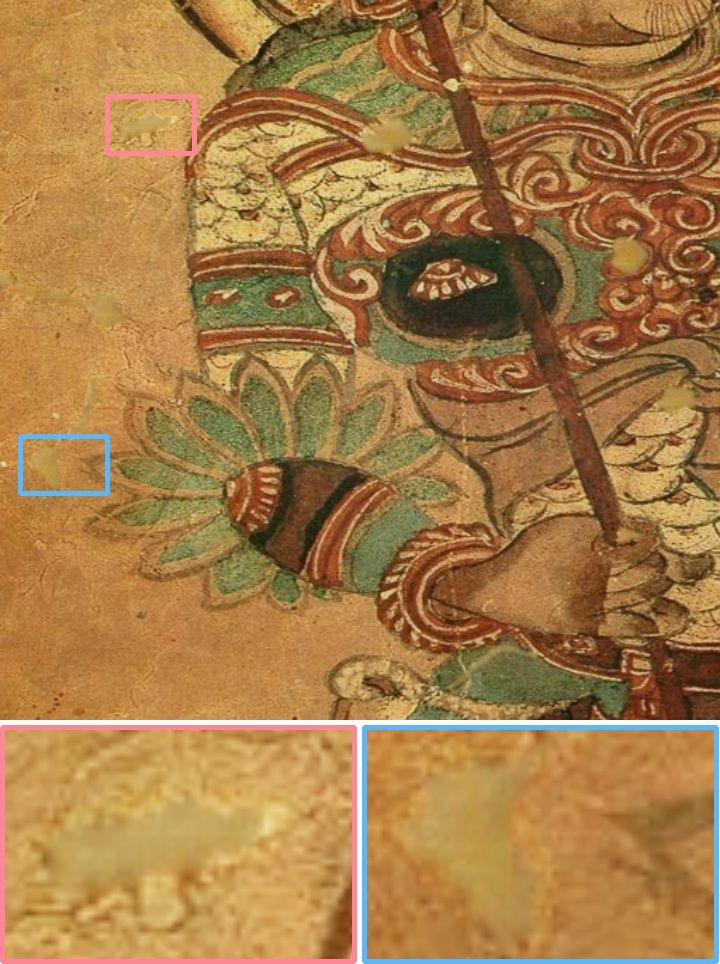}
            \centerline{\scriptsize (f) EdgeConn}
        \end{minipage}
        \hfill
        \begin{minipage}[t]{0.105\linewidth}
            \centering
            \includegraphics[width=\linewidth]{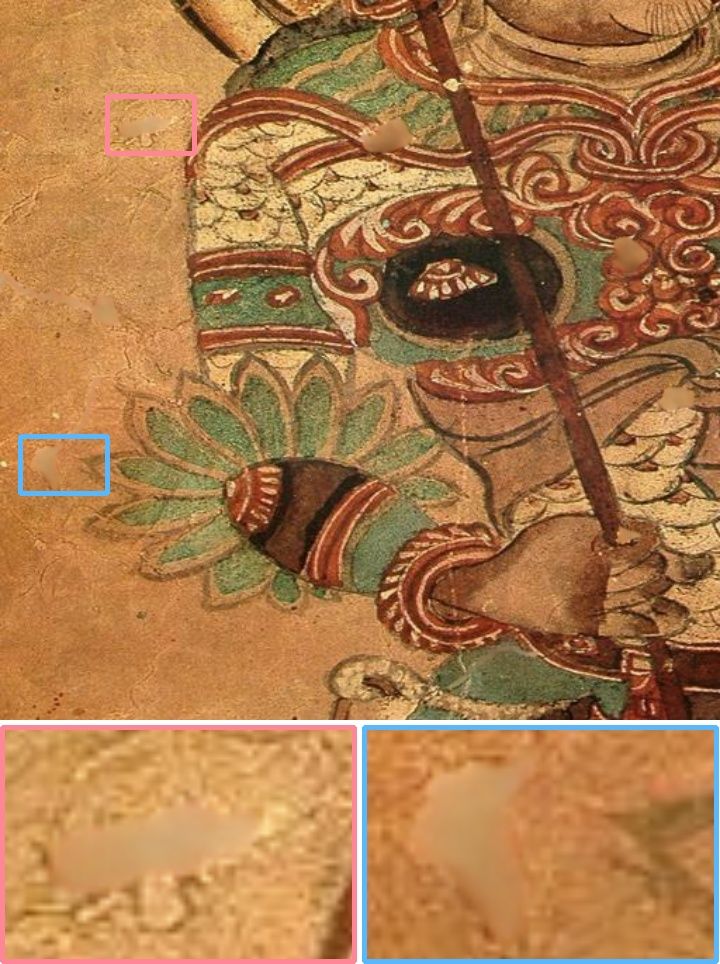}
            \centerline{\scriptsize (g) HINT}
        \end{minipage}
        \hfill
        \begin{minipage}[t]{0.105\linewidth}
            \centering
            \includegraphics[width=\linewidth]{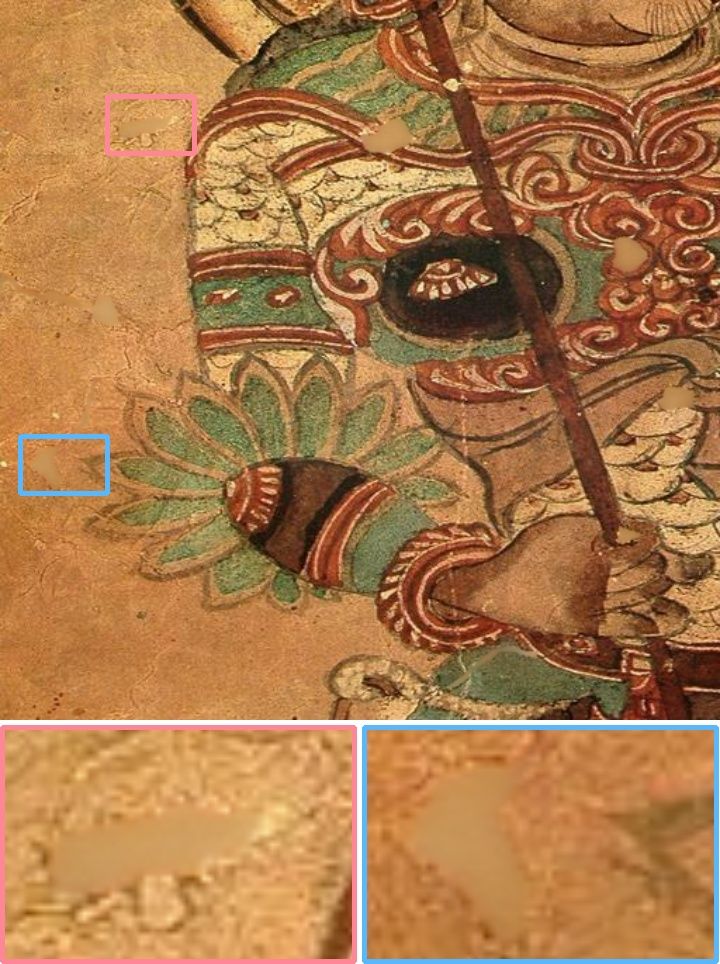}
            \centerline{\scriptsize (h) Ours}
        \end{minipage}
        \hfill
        \begin{minipage}[t]{0.105\linewidth}
            \centering
            \includegraphics[width=\linewidth]{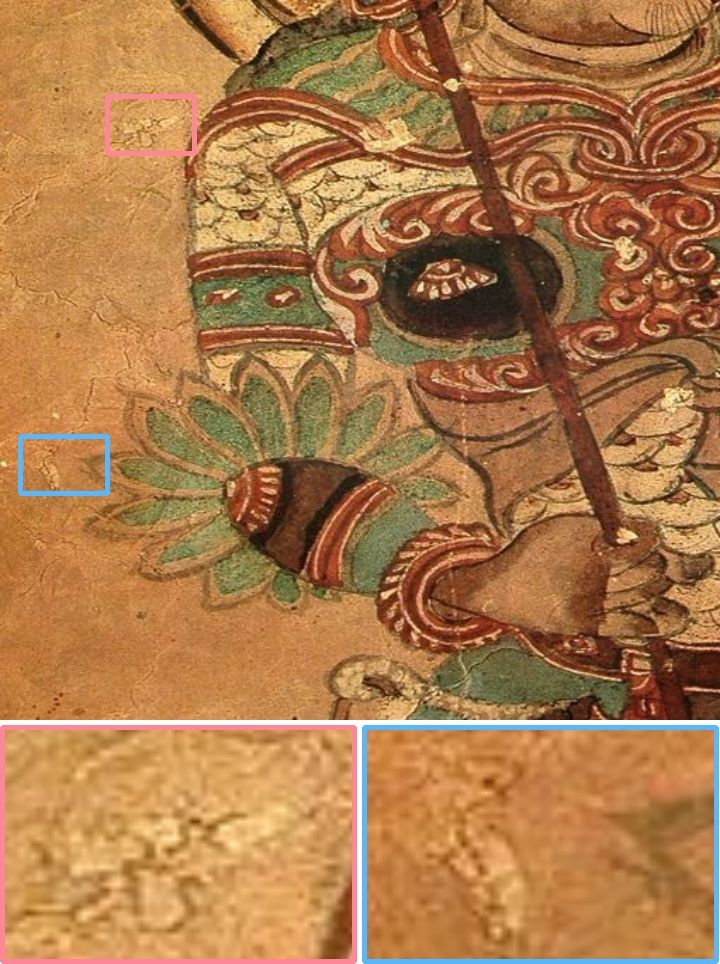}
            \centerline{\scriptsize (i) Target}
        \end{minipage}
    \end{minipage}
    
    \caption{Qualitative comparison with state-of-the-art methods. 
    From left to right: (a) Masked input, (b) CoordFill, (c) LaMa, (d) SyFormer, (e) TFormer, (f) EdgeConn (EdgeConnect abbreviated in figure labels), (g) HINT, (h) Our method, and (i) Target. Our method better preserves the structural integrity and fine details of the murals while generating more coherent and visually pleasing results.
    }
    \label{fig:compare}
\end{figure}
\subsection{Dataset and Evaluation Metrics}
To evaluate our model and compare it with other methods, we conduct experiments on the following datasets: (1) \textbf{MuralDH:} A Dunhuang mural collection with 961 high-resolution images (760 for training, 201 for testing) containing pixel-level damage annotations and deterioration masks. (2) \textbf{Dunhuang:} The Dunhuang Grottoes dataset consists of 600 mural images (500 for training, 100 for testing) with corresponding deterioration masks.

We employ multiple complementary metrics for comprehensive performance assessment: Peak Signal-to-Noise Ratio (PSNR) for overall image quality, Structural Similarity Index (SSIM)~\cite{wang2004image} for structural fidelity, Mean Absolute Error (MAE) for pixel-level accuracy, and Learned Perceptual Image Patch Similarity (LPIPS)~\cite{zhang2018unreasonable} for perceptual quality. Superior performance is indicated by higher PSNR/SSIM values and lower MAE/LPIPS scores.
\subsection{Implementation Details}
We implement our model in PyTorch and train it on 2 NVIDIA A800 GPUs. The training process uses $256\times256$ input images over 200 epochs with batch size 4, employing AdamW optimizer~\cite{loshchilov2018decoupled} (initial learning rate $2\times10^{-4}$) and Cosine Annealing scheduler~\cite{loshchilov2017sgdr}. To enhance model robustness, we apply standard data augmentation techniques including random cropping, flipping, rotation, and mix-up. For testing, images are processed at $512\times512$ resolution.
\subsection{Comparisons with State-of-the-arts}
We evaluate our proposed CMAMRNet against twelve state-of-the-art methods on both MuralDH and Dunhuang datasets from quantitative and qualitative perspectives.

\textbf{Quantitative Analysis:} As shown in Table~\ref{table:Compare}, our CMAMRNet significantly surpasses existing approaches. Compared to the previous state-of-the-art HINT~\cite{chen2024hint}, our method achieves 2.18\% and 0.21\% improvements in PSNR and SSIM on MuralDH dataset, while showing even larger gains of 2.58\% and 2.86\% on Dunhuang dataset, demonstrating superior capability in both pixel-level accuracy and structural preservation. The consistent performance improvement across different evaluation metrics highlights the effectiveness of our comprehensive mask guidance approach, particularly in preserving fine artistic details that are critical for cultural heritage conservation.

\textbf{Qualitative Analysis:} Visual comparisons in~\cref{fig:compare} further validate our method's effectiveness. While perfect ground truth restoration of degraded murals remains challenging, CMAMRNet demonstrates stronger ability in preserving authentic textures and structural details. Compared to existing approaches that often produce blurry or inconsistent results, our method generates restorations that better maintain the original mural characteristics and integrity. Specifically, our approach handles intricate crack patterns and complex degradations more effectively, producing more visually coherent and historically authentic restorations where other methods struggle with maintaining color fidelity or structural continuity at damage boundaries.
\subsection{Ablation Studies}
\begin{table}[b]
\centering
\caption{Quantitative results of ablation studies on the MuralDH dataset. \ding{51}and \ding{55} indicate the presence and absence of each component respectively.
}
\label{tab:ablation}
\adjustbox{width=0.6\linewidth}{
\begin{tabular}{ccc|cccc}
\toprule
\multirow{2}{*}{Index} & \multirow{2}{*}{MAUDS} & \multirow{2}{*}{CFA} & \multicolumn{4}{c}{Metric} \\ 
\cmidrule(r){4-7} 
& & & PSNR $\uparrow$ & SSIM $\uparrow$ & MAE $\downarrow$ & LPIPS $\downarrow$ \\ 
\midrule
(1) & \ding{55} & \ding{51} & 35.4464 & 0.9667 & 1.3852 & 0.0514 \\

(2) & \ding{51} & \ding{55} & 35.9325 & 0.9672 & 1.2004 & 0.0530 \\

(3) & \ding{51} & \ding{51} & 36.2565 & 0.9683 & 1.1454 & 0.0497 \\ 
\bottomrule
\end{tabular}
}
\end{table}
To verify the effectiveness of our key components, we conduct ablation studies on the MuralDH dataset. As shown in~\cref{tab:ablation}, without MAUDS, the model shows significant degradation in restoration quality, indicating its crucial role in maintaining mask sensitivity across scales. This degradation occurs because without MAUDS, mask information becomes diluted through feature transitions, leading to reduced restoration precision in damaged regions.

Without CFA, although MAUDS maintains basic mask awareness, the higher LPIPS value and compromised restoration results indicate the model struggles to simultaneously capture fine textures and global structures, particularly in regions requiring complex synthesis or structural continuation.

By integrating both modules, our complete model achieves superior performance in all metrics, demonstrating that MAUDS and CFA work complementarily to enable effective mask-aware restoration. This synergy allows the network to maintain precise mask guidance while achieving comprehensive feature representation at multiple scales.
\section{Conclusion}
In this paper, we present CMAMRNet, a novel framework for digital mural restoration that maintains comprehensive mask guidance throughout the processing pipeline while effectively capturing rich contextual information at multiple scales. Our key innovations include the Mask-Aware Up/Down-Sampler (MAUDS), which preserves mask sensitivity across resolution transitions through dedicated channel operations, and the Co-Feature Aggregator (CFA), which operates at both highest and lowest resolutions to extract complementary features for fine-grained details and global structures. Extensive experiments on MuralDH and Dunhuang datasets demonstrate that CMAMRNet significantly outperforms state-of-the-art methods across multiple metrics, achieving superior restoration quality while maintaining the authentic artistic elements and cultural significance of murals.

\section{Acknowledgement}
This work was supported in part by the Science and Technology Development Fund, Macau SAR, under  Grant 0141/2023/RIA2 and 0193/2023/RIA3, and the University of Macau under Grant MYRG-GRG2024-00065-FST-UMDF, in part by the Shenzhen Polytechnic University Research Fund (Grant No. 6025310023K).

\bibliography{egbib}
\end{document}